\documentclass[runningheads]{llncs}

\PassOptionsToPackage{table}{xcolor}

\usepackage{eccv}

\usepackage{eccvabbrv}
\usepackage{algorithm}
\usepackage{algorithmic}
\newcommand{\INPUT}{\item[\textbf{Input:}] }
\newcommand{\OUTPUT}{\item[\textbf{Output:}] }
\renewcommand{\algorithmicrequire}{\textbf{Input:}}
\renewcommand{\algorithmicensure}{\textbf{Output:}}
\usepackage{graphicx}
\usepackage{booktabs}
\usepackage{multirow} % 解决你现在的报错
\usepackage{siunitx} % 用于对齐数字，让表格非常整齐
\usepackage{pifont}
\usepackage{array}    % 用于控制列宽
\newcommand{\cmark}{\ding{51}}%
\newcommand{\xmark}{\ding{55}}%

\usepackage{bm}
\usepackage{enumitem}
\usepackage[accsupp]{axessibility}  % Improves PDF readability for those with disabilities.
\usepackage{xcolor}

\usepackage[leftcaption]{sidecap} 
\usepackage{wrapfig}

\usepackage{marvosym}

\usepackage{hyperref}

\usepackage{orcidlink}

\makeatletter
\def\thanks#1{\protected@xdef\@thanks{\@thanks
        \protect\footnotetext{#1}}}
        
\begin{document}

%\title{UnitDiff: Unified Diffusion Decoder with Temporal Transition Modulation for End-to-End Driving} 

\title{
UniTeD: Unified Temporal Diffusion for Joint Perception and Planning in Autonomous Driving
}

\titlerunning{UniTeD for Autonomous Driving}

% Second Author\inst{2,3}\orcidlink{1111-2222-3333-4444} \and
% Third Author\inst{3}\orcidlink{2222--3333-4444-5555}}

% \author{
% Bo Zhao\inst{1}\thanks{
% \inst{\star} Equal contribution.} \and
% Xinting Zhao\inst{1}\inst{\star} \and
% Naifan Li\inst{1} \and
% Erkang Cheng\inst{1}\textsuperscript{\Letter}
% \thanks{
% \textsuperscript{\Letter} Corresponding author.
% } \and
% Haibin Ling\inst{2}
% }
% \institute{
% \inst{1} Nullmax \\
% \inst{2} Westlake University
% }

\author{
Bo Zhao\inst{1}\textsuperscript{$\star$}\orcidlink{0009-0006-5861-9692} \thanks{
$^\star$ Equal contribution.} \and
Xinting Zhao\inst{1}\textsuperscript{$\star$}\orcidlink{0009-0007-0602-378X}\and
Naifan Li\inst{1}\orcidlink{0009-0004-4147-3151} \and
Erkang Cheng\inst{1}\textsuperscript{\Letter}\orcidlink{0000-0001-7941-6911}
\thanks{
\textsuperscript{\Letter} Corresponding author.
}
\and
Haibin Ling\inst{2}\orcidlink{0000-0003-4094-8413}
}
\institute{
Nullmax \and
Westlake University \\
%\email{\{fullname\}@nullmax.ai, linghaibin@westlake.edu.cn}
\email{\{zhaobo,zhaoxinting,linaifan,chengerkang\}@nullmax.ai, linghaibin@westlake.edu.cn}
}

\authorrunning{B.~Zhao et al.}

% Springer Heidelberg, Tiergartenstr.~17, 69121 Heidelberg, Germany

% \institute{
% Nullmax \\
% \email{\{tangyingqi,mengzhaotie,chenguoliang,chengerkang\}@nullmax.ai}
% }

\maketitle

\begin{abstract}

Diffusion models have shown strong potential for multi-modal planning in end-to-end autonomous driving. However, most existing methods confine diffusion to the planning module, conditioning on fixed outputs from separate discriminative perception networks. This decoupled design propagates perception errors to the planner, increasing optimization difficulty and reducing robustness. 
To overcome these limitations, we propose \textit{UniTeD}, a \textit{Uni}fied \textit{Te}mporal \textit{D}iffusion framework that jointly models perception and planning through iterative denoising in a shared generative space. 
%The proposed framework also enables bidirectional information exchange between the two tasks and promotes mutual refinement and enhances robustness through noise-conditioned multi-task training.
By enabling bidirectional information exchange, the framework facilitates mutual refinement between tasks and improves robustness via noise-conditioned multi-task training.
We further extend this unified diffusion paradigm to a streaming setting by incorporating temporal context. A Temporal Transition Module (TTM) is introduced to resolve the noise-level mismatch between historical and current frames.
In addition, we propose an Anchor Refresh Strategy (ARS) to alleviate the training–inference distribution shift commonly observed in sparse diffusion-based end-to-end driving frameworks.
Without bells and whistles, UniTeD achieves state-of-the-art performance across multiple benchmarks, surpassing both recent discriminative end-to-end methods and diffusion-based planning approaches. 
%The code will be publicly released upon acceptance of the paper.
  
\end{abstract}
\section{Introduction}
\label{sec:intro}

Autonomous driving is undergoing a fundamental paradigm shift, moving from traditional modular pipelines, with perception~\cite{huang2021bevdet, li2024bevformer, lin2022sparse4d, liu2022petr, li2022hdmapnet, liao2022maptr, liao2025maptrv2}, motion prediction~\cite{gao2020vectornet, gu2023vip3d, jiang2022perceive, zeng2022motr}, and planning~\cite{cheng2024pluto, cui2021lookout, sadat2020perceive, dauner2023parting} as separate components, toward fully \textit{end-to-end} (E2E) systems~\cite{hu2023planning, jiang2023vad, chitta2022transfuser, liao2025diffusiondrive}. E2E approaches reformulate autonomous driving as a fully differentiable process that directly optimizes the final planning objective. %This paradigm enables simpler deployment and reduces explicit error propagation through streamlined and lossless information flow.

\begin{figure}[t]
    \centering
    \includegraphics[width=0.90\linewidth, keepaspectratio]{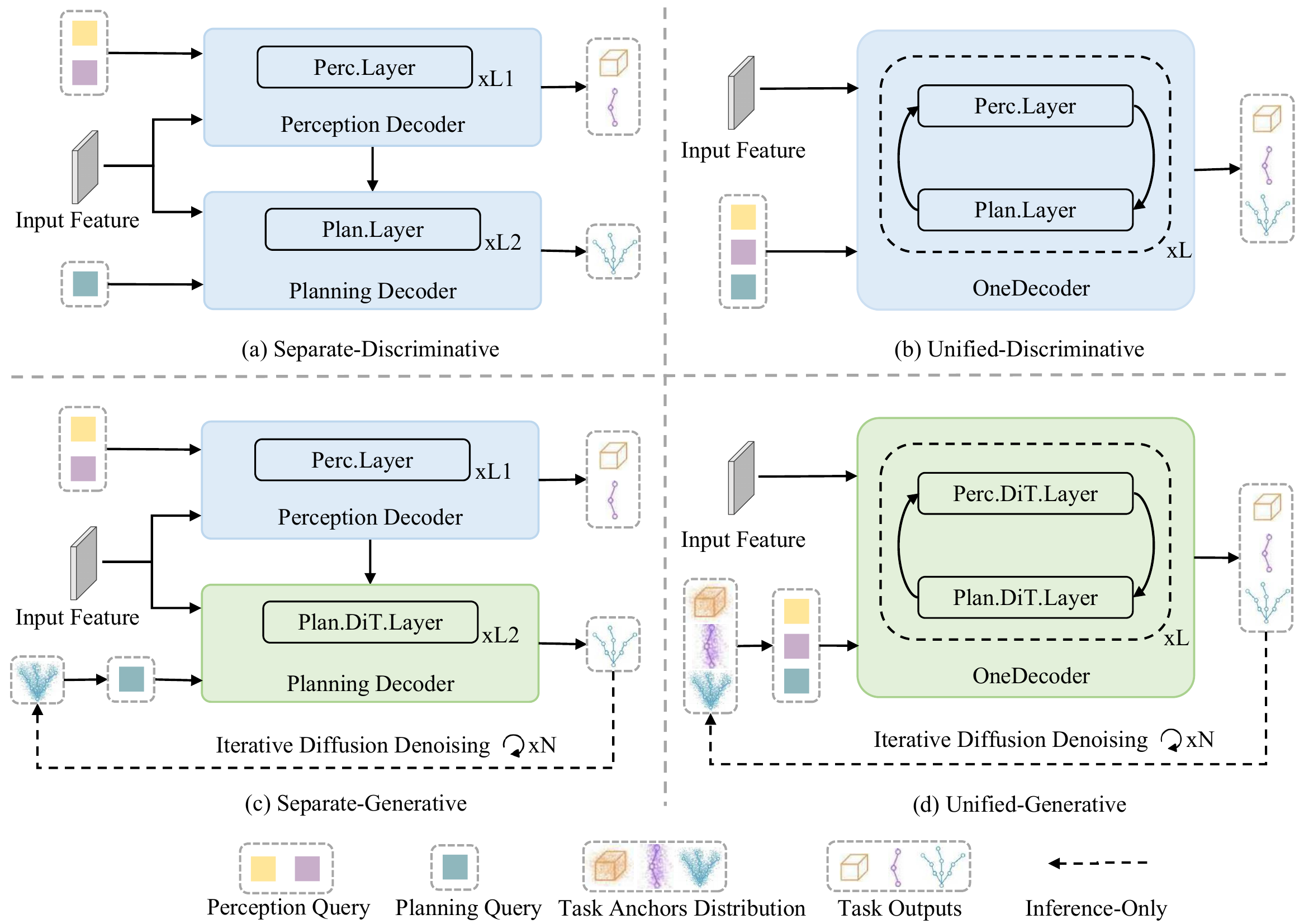}
    \caption{Comparison of existing paradigms for end-to-end autonomous driving. (a) Separate-Discriminative: separate perception and planning with discriminative modeling for both tasks. (b) Unified-Discriminative: unified perception and planning with discriminative modeling for both tasks. (c) Separate-Generative: separate perception and planning with generative modeling for planning task only. (d) Unified-Generative (ours): unified perception and planning with generative modeling for both tasks.
}
    \label{fig:figure1}
\end{figure}

Most recent E2E approaches adopt a discriminative formulation~\cite{hu2023planning, jiang2023vad, sun2025sparsedrive, chitta2022transfuser, weng2024drive, jia2025drivetransformer, tang2025hip}. They typically treat perception and planning separately (Fig.~\ref{fig:figure1} (a)), either sequentially~\cite{hu2023planning, jiang2023vad, sun2025sparsedrive} or in parallel~\cite{chitta2022transfuser, weng2024drive}. For instance, UniAD~\cite{hu2023planning} and VAD~\cite{jiang2023vad} employ sequential pipelines, where perception outputs are passed to a downstream planning module. In contrast, TransFuser~\cite{chitta2022transfuser} and PARA-Drive~\cite{weng2024drive} process perception and planning tasks concurrently, but with limited cross-task interaction. To enable deeper integration, DriveTransformer~\cite{jia2025drivetransformer} and HiP-AD~\cite{tang2025hip} introduce unified decoders that jointly process all task queries. As shown in Fig.~\ref{fig:figure1} (b), perception (agent and map) and planning queries interact at every layer and directly attend to shared feature representations, substantially improving E2E driving performance. 
However, these discriminative methods share a core limitation rooted in their optimization objective: minimizing supervised loss encourages predictions toward mean or majority behaviors, limiting their ability to capture the multi-modal nature of driving.

To address this issue, recent research has explored generative modeling, particularly diffusion-based approaches~\cite{chi2025diffusion, liao2025diffusiondrive, zheng2025resad, yin2025diffrefiner} (Fig.~\ref{fig:figure1} (c)). DiffusionDrive~\cite{liao2025diffusiondrive} proposes a truncated diffusion policy that stabilizes generation while reducing denoising steps for real-time efficiency. ResAD~\cite{zheng2025resad} addresses spatiotemporal imbalance by reformulating trajectory prediction as residual learning. DiffRefiner~\cite{yin2025diffrefiner} further introduces a two-stage framework, where a diffusion-based refiner denoises coarse trajectories generated in an initial stage.
Despite advances achieved, existing generative frameworks limit diffusion modeling to the planning task, treating perception outputs as fixed conditions. This strategy may propagate perception errors to the generative process, leading to inaccurate guidance and increased optimization complexity. It thus prevents joint refinement between perception and planning, restricting the full potential of generative modeling. % in E2E autonomous driving.

Furthermore, existing approaches use only single-frame information for diffusion-based planning and ignore temporal dynamics. %, limiting their planning performance.
In addition, sparse query-based diffusion planning methods~\cite{liao2025diffusiondrive, yin2025diffrefiner} often suffer from severe training-inference inconsistency: during training, only a small subset of planning queries matched to the ground truth are optimized, while the remaining queries receive little supervision, causing a query distribution shift at inference time. 
Such behavior fundamentally contradicts the iterative refinement principle of diffusion models, where outputs are expected to progressively improve as denoising proceeds.

To address the aforementioned challenges, we present \textit{UniTeD}, a \textit{Uni}fied \textit{Te}mporal \textit{D}iffusion framework that jointly models perception and planning through iterative denoising in a shared generative space. Specifically, UniTeD adopts a unified diffusion decoder that jointly processes perception queries (agents and map elements) and planning queries within a single generative process. By denoising them simultaneously in a shared generative space, our approach fully leverages the strong uncertainty modeling capability of diffusion models and enables comprehensive cross-task information exchange. 
%Specifically, our approach adopts a unified diffusion decoder that jointly takes perception queries (agents and map elements) and planning queries as input and denoises them simultaneously, enabling comprehensive interaction between perception and planning within a single generative process. This design facilitates deep cross-task information exchange in a shared generative space, while fully leveraging the strong uncertainty modeling capability of diffusion models. 
In this way, UniTeD not only supports multi-modal trajectory generation for the ego vehicle but also strengthens perception tasks, including dynamic agent prediction and static map understanding.
Importantly, optimization under a noise-conditioned diffusion paradigm inherently improves multi-task robustness. Each task is trained to ensure accurate predictions even when conditioned on noisy or imperfect intermediate outputs from other tasks, leading to greater stability and improved generalization at inference time.

Additionally, to better capture temporal dynamics that are often overlooked in existing diffusion-based planners, we incorporate a memory bank that stores historical task queries, enabling a streaming unified diffusion framework. We introduce a Temporal Transition Module (TTM) to solve the noise-level mismatch between historical and current frames.
Furthermore, we propose an Anchor Refresh strategy (ARS) to mitigate the training–inference distribution shift in sparse diffusion frameworks. During inference, high-confidence queries are retained, while low-confidence task queries are refreshed by resampling from the original noise distribution. This mechanism preserves query diversity while maintaining alignment with the training distribution, ensuring stable and reliable performance even in complex multi-modal settings.

Extensive experiments demonstrate that UniTeD achieves strong performance across major benchmarks, outperforming both discriminative E2E approaches and recent diffusion-based planning ones. In particular, our method attains 87.25 DS on Bench2Drive, as well as 90.24 PDMS and 90.13 EPDMS on NAVSIM. 
%These results verify that both perception and planning benefit substantially from the proposed unified and streaming diffusion framework.

The main contributions of this work are summarized as follows:
\begin{enumerate}[label=\arabic*.]
\item We propose a novel unified diffusion framework for E2E autonomous driving. Our solution jointly denoises perception and planning queries, enabling deep bidirectional information exchange within a shared generative space.
\item We extend unified diffusion to a streaming setting via a memory bank and introduce a Temporal Transition Module to resolve noise-level mismatches between historical and current frames.
\item We present an Anchor Refresh Strategy to alleviate training–inference distribution shifts in sparse diffusion-based autonomous driving frameworks.
\item We show superior performance across major benchmarks, surpassing both state-of-the-art discriminative E2E methods and  diffusion-based planners.
\end{enumerate}

\section{Related Works}
%\subsection{Discriminative End-to-End Autonomous Driving}
\textbf{Discriminative End-to-End Autonomous Driving.~}  
%The evolution of End-to-End autonomous driving has transitioned from modular pipeline, where perception~\cite{huang2021bevdet, li2024bevformer, lin2022sparse4d, liu2022petr, li2022hdmapnet, liao2022maptr, liao2025maptrv2}, prediction~\cite{gao2020vectornet, gu2023vip3d, jiang2022perceive, zeng2022motr}, and planning~\cite{cui2021lookout, sadat2020perceive, dauner2023parting} were cascaded sequentially, toward single architecture. 
Most recent E2E approaches typically apply discriminative methods for both the planning and perception tasks.
UniAD~\cite{hu2023planning} represents a milestone in this category, integrating multiple tasks into a single model, while VAD~\cite{jiang2023vad} further improves efficiency by replacing dense rasterized scene representations with vectorized ones. SparseDrive~\cite{sun2025sparsedrive} eliminates the dependency on dense BEV features through a fully sparse query-based framework. To mitigate the sequential dependencies, several works explore parallel architectures. TransFuser~\cite{chitta2022transfuser} fuses LiDAR and camera inputs to construct a BEV representation, executing perception and planning tasks concurrently. PARA-Drive~\cite{weng2024drive} extends this by jointly modeling mapping, motion, and occupancy prediction alongside planning. DriveTransformer~\cite{jia2025drivetransformer} introduces a unified decoder that jointly processes all task queries. 
HiP-AD~\cite{tang2025hip} similarly employs a unified decoder for both perception and planning tasks, and further enhances performance by incorporating multi-granularity planning queries.
%\hlb{Unlike these methods, which treat both perception and planning as discriminative tasks, our UniTeD formulates them as generative problems and employs a unified decoder to perform them jointly.}
%HiP-AD~\cite{tang2025hip} incorporates a multi-granularity planning query representation to provide additional supervision for trajectory prediction. Another branch of research attempts to enhance multi-modal capabilities via anchor-based classification methods. VADv2~\cite{chen2024vadv2} scores and samples from a massive fixed vocabulary of anchor trajectories. Hydra-MDP~\cite{li2024hydra} and Hydra-MDP++~\cite{li2025hydra} utilize rule-based scorers to build richer supervisory signals. 
%\hlb{Different from these discriminative frameworks that often rely on independent regression or classification heads, our UniTeD reformulates E2E driving as a joint generative process. This allows perception and planning to benefit from bidirectional refinement within a shared latent space, effectively mitigating the error propagation typical of decoupled discriminative pipelines while providing the inherent multi-modal generative capability to handle the complex uncertainty of driving scenarios.}
%\subsection{Generative Methods in Autonomous Driving}

\noindent\textbf{Generative Methods in Autonomous Driving.~} 
In the perception domain of autonomous Driving, diffusion-based frameworks have redefined traditional tasks as a denoising process. DiffusionDet~\cite{chen2023diffusiondet} first redefines object detection as a denoising diffusion process from noisy boxes to object boxes. MotionDiffuser~\cite{jiang2023motiondiffuser} utilizes diffusion processes to model the multi-modal distribution of multi-agent motion prediction. DiffusionTrack~\cite{luo2024diffusiontrack} introduces a noise-to-tracking paradigm, reformulating multi-object tracking as a conditional denoising task to achieve more robust data association. DiffMap~\cite{jia2024diffmap} utilizes generative models for map segmentation. PolyDiffuse~\cite{chen2023polydiffuse} further advances this by generating vectorized map elements through a denoising process. LaneDiffusion~\cite{wang2025lanediffusion} strengthens the representation of complex road networks by effectively modeling lane centerlines and topological connections.

In the planning domain, DiffusionDrive~\cite{liao2025diffusiondrive} proposes a truncated diffusion policy to stabilize generation while enhancing real-time efficiency through reduced denoising steps. ResAD~\cite{zheng2025resad} reformulates trajectory prediction as residual learning to address spatiotemporal imbalances. DiffRefiner~\cite{yin2025diffrefiner} employs a two-stage framework, utilizing a diffusion-based refiner to polish coarse trajectories from the initial stage. DiffusionDriveV2~\cite{zou2025diffusiondrivev2} incorporates reinforcement learning to prune low-quality modes and explore superior trajectories. DiffAD~\cite{wang2025diffad} reformulates autonomous driving as a conditional image generation task, predicting future scenes in the form of rasterized BEV images.
% \hlb{
% These methods retain diffusion-based models for planning while treating discriminative perception outputs as fixed conditions for the planner. In contrast, our approach jointly formulates perception and planning within a shared generative space. Furthermore, we extend the framework to a streaming scheme.
% }

Being a generative solution, our UniTeD jointly formulates perception and planning within a shared generative space, while previous methods retain diffusion-based models for planning while treating discriminative perception outputs as fixed conditions for the planner. Furthermore, UniTeD extends the generative framework to a streaming scheme.

% \hlb{However, most existing methods still follow a cascaded generative paradigm, either focusing on independent perception tasks or treating perception as a static condition for a separate diffusion-based planner. This unidirectional dependency limits the synergy between scene understanding and action generation. UniTeD breaks this bottleneck by reformulating autonomous driving as a joint denoising problem. By unifying perception and planning in a shared generative space, UniTeD enables bidirectional refinement, ensuring that the final outputs are not only individually accurate but also holistically consistent and physically plausible.}

%\subsection{Temporal Modeling in Autonomous Driving}

\noindent\textbf{Temporal Modeling in Autonomous Driving.~} 
Existing E2E frameworks typically incorporate temporal dynamics through two main streams: Global Feature Fusion~\cite{hu2023planning,jiang2023vad} and Query-based Propagation~\cite{sun2025sparsedrive,jia2025drivetransformer,tang2025hip,song2025don}. Early methods like UniAD~\cite{hu2023planning} and VAD~\cite{jiang2023vad} utilize stacked historical BEV features to provide a global spatial-temporal context for downstream tasks. SparseDrive~\cite{sun2025sparsedrive}, DriveTransformer~\cite{jia2025drivetransformer}, and HiP-AD~\cite{tang2025hip} all utilize a memory queue to cache historical queries, leveraging temporal attention to integrate historical information. To ensure planning stability, MomAD~\cite{song2025don} further introduces a momentum mechanism to select the optimal planning query, which is then integrated with historical queries via cross-attention. Despite the success of these discriminative frameworks, incorporating temporal modeling into diffusion-based E2E frameworks remains challenging. Existing diffusion planners, such as DiffusionDrive~\cite{liao2025diffusiondrive} and ResAD~\cite{zheng2025resad}, primarily focus on single-frame generation or naive concatenation of historical features. 
Different from these methods, UniTeD introduces a unified diffusion-based streaming paradigm that propagates joint task queries across frames. 
A novel module is also proposed to solve the noise-level mismatch between historical and current frames.
%Unlike discriminative methods, where historical features are inherently clean and directly reusable, our streaming framework introduces a unique noise-level mismatch between historical memory and the current noisy queries.

%Unlike discriminative methods where historical features are inherently ``clean'', our streaming approach encounters a unique noise-level mismatch between historical memory and current noisy queries. To bridge this gap, UniTeD utilizes a novel Temporal Transition Module (TTM) to explicitly align the noise manifolds across different timestamps, ensuring that historical priors are distributionally consistent with the current noising stage for stable, long-term driving.}

%\noindent\textbf{\ling{Our method.}~} \ling{summarize the difference bw our metod and previoius ones.}
\section{Method}
\subsection{Overview}

\begin{figure*}[!t]
    \centering
    \includegraphics[width=0.95\linewidth, keepaspectratio]{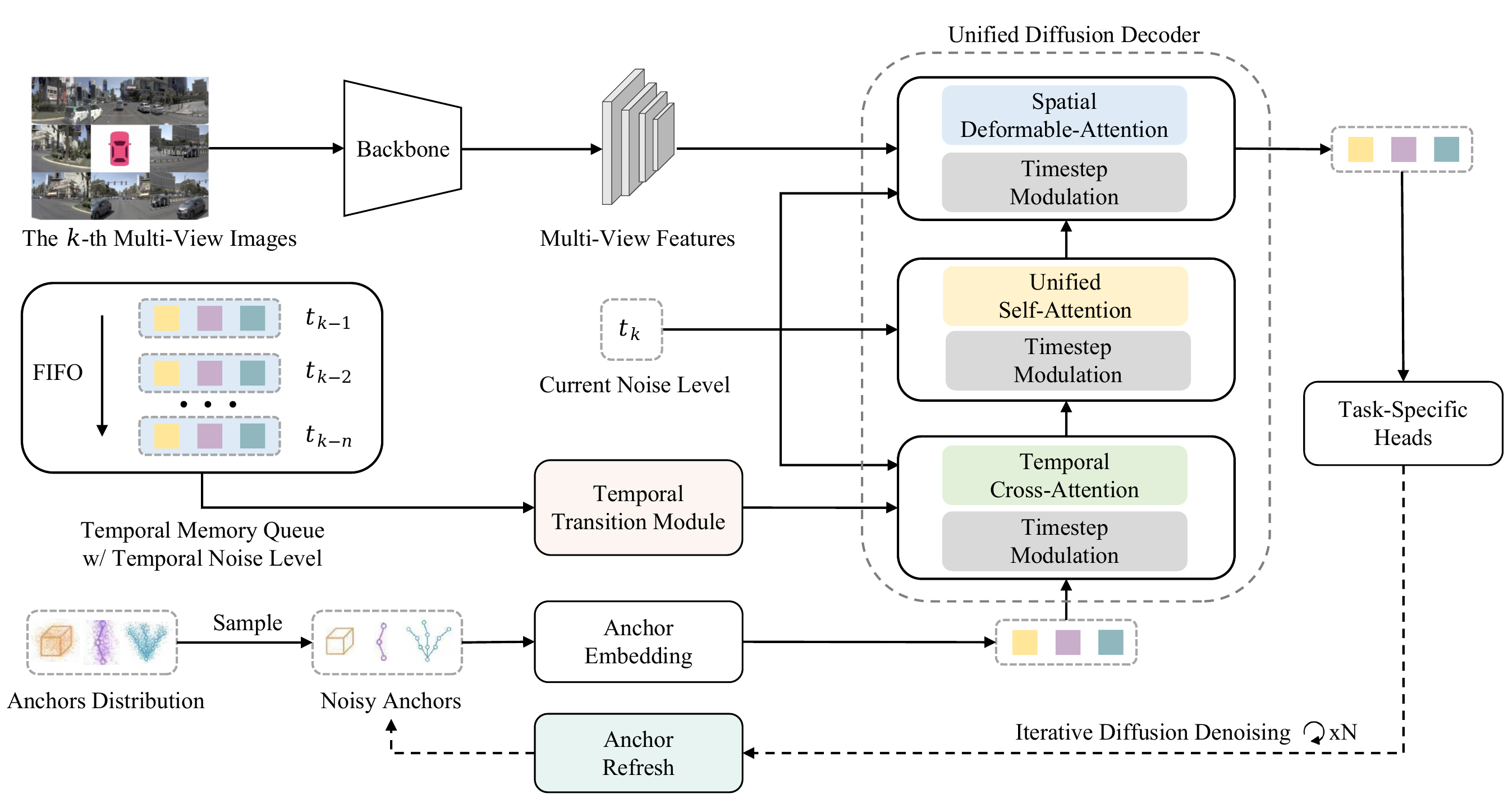}
    %\caption{Overview of UniTeD. Our framework follows the Uni-Gen paradigm for End-to-End driving. Beyond the standard Backbone, Task-Specific Heads and Anchor Embedding, we introduce three core components: (a) Unified Diffusion Decoder, which performs interaction between the joint query set, visual features, and modulated historical context to generate updated all-task queries; (b) Timestep Transition Module (TTM), which aligns historical temporal features with the current noise level to produce a consistent modulated context; and (c) Anchor Refreshing, a strategy that refresh low-confidence task queries during iterative denoising to maintain an input distribution consistent with the training stage.}
    \caption{Overview of UniTeD. %Following the unified-generative paradigm, 
    UniTeD contains three core components: (a) Unified Diffusion Decoder that models perception and planning queries through iterative denoising in a shared generative space; (b) Temporal Transition Module (TTM) that resolves the noise-level mismatch between historical and current frames; and (c) Anchor Refresh Strategy that alleviates the training–inference distribution shift. 
    %Here, $k$, $t_k$ denote the frame index, and noise level associated with $k$-th frame, respectively.
    }
    \label{fig:figure2}
\end{figure*}

The overall architecture of UniTeD is illustrated in Fig.~\ref{fig:figure2}, which operates as a unified diffusion process. First, the image backbone extracts multi-scale features $\mathbf{F} = \{\mathbf{F}_v \}_{v=1}^V$ from $V$ multi-view images at $k$-th frame. Second, we define $\mathbf{A} = \{\mathbf{A}^{a} \in \mathbb{R}^{N_{a} \times D_{a}}, \mathbf{A}^{m} \in \mathbb{R}^{N_{m} \times D_{m}}, \mathbf{A}^{p} \in \mathbb{R}^{N_{p} \times D_{p}} \}$ as a joint anchor set for agent, map, planning tasks, respectively.  $N_{a}$, $N_{m}$, and $N_{p}$ denote the numbers of anchors for the different tasks, and $D_{a}$, $D_{m}$, and $D_{p}$ denote their corresponding anchor dimensions. Following the truncated diffusion~\cite{liao2025diffusiondrive}, the noisy anchors $\mathbf{A}_{t_k}$ are sampled from a Gaussian distribution centered around the prior $\mathbf{A}$, where $t_k$ denotes the noise level associated with the $k$-th frame. Then $\mathbf{A}_{t_k}$ are projected into a shared latent query space by an Anchor Embedding module, yielding $\mathbf{Q}_{k} = \{\mathbf{Q}^{a}, \mathbf{Q}^{m}, \mathbf{Q}^{p}\} \in \mathbb{R}^{({N_{a} + N_{m} + N_{p}}) \times C}$. 
These queries are then fed into the Unified Diffusion Decoder (Sec.~\ref{sec:decoder}).
To incorporate temporal context, the Temporal Transition Module (TTM) (Sec.~\ref{sec:ttm}) modulates a temporal queue of historical features $\{\mathbf{Q}_{k-i}\}_{i=1}^n$ to generate aligned historical features $\{\tilde{\mathbf{Q}}_{k-i}\}_{i=1}^n$ consistent with the current noise level $t_k$. Together with the multi-view features $\mathbf{F}$, the modulated temporal features serve as conditional inputs to the decoder.
Finally, after denoising, task-specific heads transform the refined queries into final predictions for the agent, map, and planning tasks. 

During inference, an Anchor Refresh Strategy (Sec.~\ref{sec:ars}) updates the predicted outputs to initialize the subsequent denoising step, enabling iterative refinement until the final timestep is reached.

\subsection{Unified Modeling via Diffusion}

We follow the truncated diffusion policy in DiffusionDrive~\cite{liao2025diffusiondrive}, but extend it to a joint multi-task modeling form. The Unified Diffusion Decoder simultaneously models the distribution of all task instances by denoising the noisy anchors $\mathbf{A}_{t_k}$ into their final clean outputs. The forward and reverse processes are as follows:

\noindent\textbf{Unified Forward Diffusion Process.~~}
During the forward process, we apply the same level of noise to all tasks simultaneously. For each anchor $a \in \mathbf{A}$, the unified forward process at noise timestep $t_k$ is defined as:
\begin{equation}
    \begin{aligned}
        q(a_{t_k} | a) = \mathcal{N}(a_{t_k}; \sqrt{\bar{\alpha}_{t_k}} a, (1 - \bar{\alpha}_{t_k}) \mathbf{I}), \quad t_k \in [1, T_{\text{trunc}}]
    \end{aligned}
    \label{eq:forward_diffusion}
\end{equation}
where $T_{\text{trunc}} \ll T_{\text{max}}$ denotes the maximum number of truncated diffusion steps. The resulting noisy anchors serve as the input to the decoder. 
This synchronized noise injection ensures that all task categories are embedded with a shared uncertainty scale, promoting consistent representation learning across tasks.

\noindent\textbf{Unified Reverse Denoising Process.~~}
In the reverse process, the decoder recovers the clean unified all-task instances from the noisy anchor inputs. We follow the DDIM sampling~\cite{song2020denoising} to iteratively refine the task states. During each denoising step, for $k$-th frame, the decoder takes the current step noisy anchors $\mathbf{A}_{t_k}$, multi-view visual features $\mathbf{F}$, and modulated historical context queue $\{\tilde{\mathbf{Q}}_{k-i}\}_{i=1}^n$ as inputs to estimate the clean state of each instance. 
%By sharing a consistent noise timestep $t$ across $\mathbf{A}_{a}, \mathbf{A}_{m},$ and $\mathbf{A}_{p}$ at the current physical frame $k$, we ensure that the entire scene is treated as a synchronized holistic entity during the generative process. This allows the model to effectively resolve spatial-temporal conflicts and ensure that the final outputs are distributionally aligned and respect the underlying physical constraints of the environment.

\subsection{Unified Diffusion Decoder}
\label{sec:decoder}

The Unified Diffusion Decoder serves as the generative core of our framework (Fig.~\ref{fig:figure2}). 
To progressively recover clean task states, the decoder refines the joint query set $\mathbf{Q}_k$ through $N$ denoising iterations across $L$ stacked blocks, each consisting of the specialized interaction layers details below:

\noindent{\textbf{Conditional Modulation.}} 
To condition the denoising process on the generative stage, every interaction layer within the decoder block is governed by a Conditional Modulation module. Following the Adaptive Layer Normalization (AdaLN) design in DiT~\cite{peebles2023scalable}, the condition $\mathbf{C}_k$, by encoding current timestep $t_k$, fed into an MLP to produce modulation parameters $\{\alpha, \beta,\gamma \}$, which dynamically projects task features onto the corresponding distribution level via AdaLN.

\noindent{\textbf{Temporal Cross-Attention Layer.}} 
After the initial modulation, a Temporal Cross-Attention layer integrates the modulated historical context queue $\{\tilde{\mathbf{Q}}_{k-i}\}_{i=1}^n$, which is aligned by the TTM to ensure noise-level consistency. The complete temporal interaction pipeline are elaborated in Sec.~\ref{sec:ttm}.

\noindent{\textbf{Unified Self-Attention Layer.}}
To enable cross-task reasoning, the joint queries at $k$-th frame $\mathbf{Q}_k = \{ \mathbf{Q}^{a}, \mathbf{Q}^{m}, \mathbf{Q}^{p} \}$ are fed into a Unified Self-Attention Layer, allowing each token to attend to all others regardless of task origin (agent, map, or planning).
This all-to-all interaction jointly updates agent motion, map geometry and ego-planning states, enabling the model to exploit cross-task geometric priors during generation.
For example, ego planning $\mathbf{Q}^{p}$ and agent intentions $\mathbf{Q}^{a}$ impose spatial constraints that assist map reconstruction $\mathbf{Q}^{m}$ in occluded regions, while map topology regularizes multi-modal trajectory generation. A single joint attention pass thus promotes globally plausible, physically and logically consistent scene synthesis.

\noindent{\textbf{Spatial Deformable Attention layer.}}
To anchor latent features in 3D space, each query $q \in \mathbf{Q}_k$ is linked to a 3D anchor $a \in \mathbf{A}$, which is projected onto the $v$-th camera image plane via the view transformation $\mathcal{P}_v$. We then apply the Spatial Deformable Attention, implemented via Multi-Scale Deformable Attention~\cite{zhu2020deformable}, to efficiently sample and aggregate multi-view, multi-scale image features $\mathbf{F}$. 

\subsection{Temporal Interaction via Temporal Transition Module}
\label{sec:ttm}

% \begin{figure}
%     \centering
%     \includegraphics[width=0.9\linewidth]{fig/fig3_byf_v1.pdf}
%     \caption{Enter Caption}
%     \label{fig:figure3}
% \end{figure}

% \begin{figure}[t]
%     \centering
%     \includegraphics[width=0.7\linewidth, keepaspectratio]{fig/fig3_final.pdf}
%     %\includegraphics[width=0.7\linewidth, keepaspectratio]{fig/fig3_bz.pdf}
%     \caption{Architecture of the Temporal Interaction via Timestep Transition Module. Our temporal interaction enables a streaming unified diffusion framework by aligning historical queries with the current noising level, and consists of three key operations: (Left) TTM, which employs historical conditions to modulate historical queries, effectively resolving the noise-level mismatch; (Right) Conditional Modulation, which injects the current condition into the current query through AdaLN; and (Top) Temporal Fusion, which aggregates the modulated historical features via cross-attention to update the current query with temporal context.}
%     \label{fig:figure3}
% \end{figure}

%\begin{wrapfigure}{r}{0.66\textwidth}    
\begin{figure}[!t]
    \centering
    \includegraphics[width=.7\linewidth, keepaspectratio]{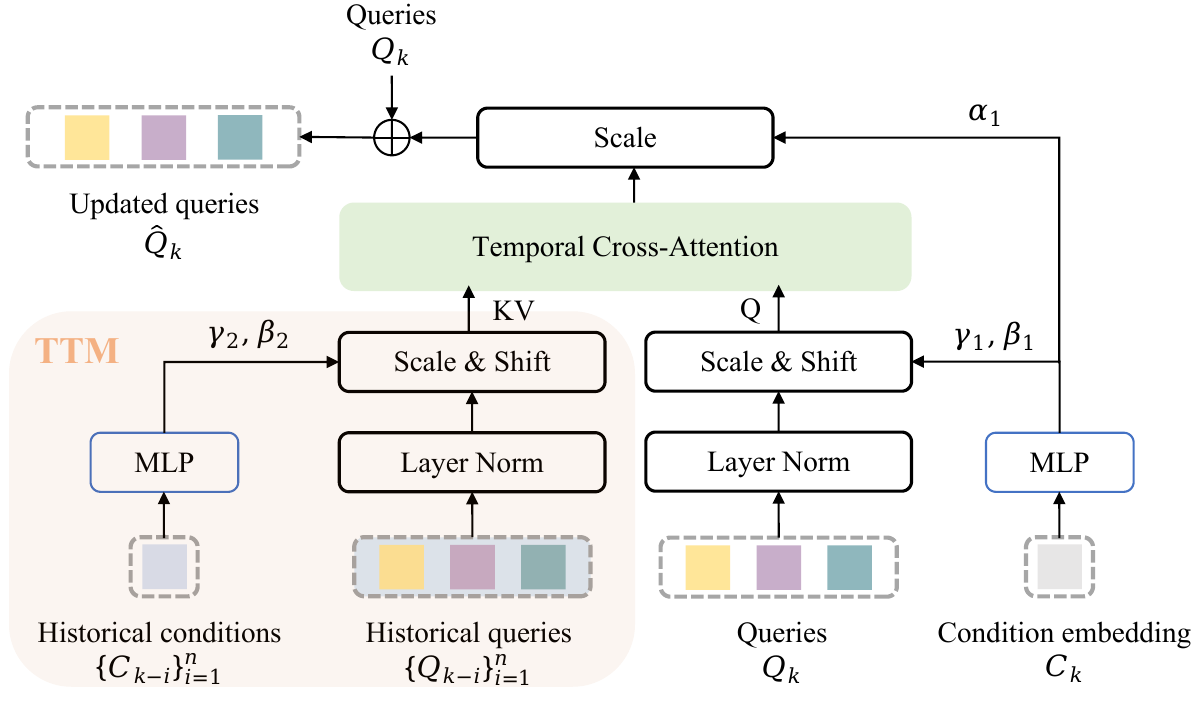}
    %\includegraphics[width=0.7\linewidth, keepaspectratio]{fig/fig3_bz.pdf}
    %\caption{Temporal Interaction via Timestep Transition Module. Our temporal interaction enables a streaming unified diffusion framework by aligning historical queries with the current noising level, and consists of three key operations: (Left) TTM, which employs historical conditions to modulate historical queries, effectively resolving the noise-level mismatch; (Right) Conditional Modulation, which injects the current condition into the current query through AdaLN; and (Top) Temporal Fusion, which aggregates the modulated historical features via cross-attention to update the current query with temporal context.}
    \caption{%Temporal Interaction via 
    Temporal Interaction. The core TTM enables a streaming unified diffusion framework by aligning historical queries with the current noising level.}
    \label{fig:figure3}
\end{figure}
%\end{wrapfigure}

The Temporal Interaction via Temporal Transition Module (TTM) (Fig.~\ref{fig:figure3}) is the key component enabling our streaming unified diffusion framework. While many diffusion-based models neglect temporal dynamics, we use a memory bank for long-range reasoning. A naive fusion of historical features, however, suffers from noise-level mismatch between historical queries $\{\mathbf{Q}_{k-i}\}_{i=1}^n$ and current queries $\mathbf{Q}_k$, as their denoising time steps are sampled stochastically and independently. Addressing this issue, we implement a structured interaction pipeline:

\noindent{\textbf{Input.}}
The four inputs at the $k$-th frame with noise level $t_k$ are:
\begin{enumerate}[label=\textbullet]
%\begin{itemize}
\item Queries $\mathbf{Q}_k$: joint noisy queries targeted for denoising at noise level $t_k$.
\item Condition embedding $\mathbf{C}_k$: $\mathbf{C}_k = \texttt{MLP}(\texttt{TE}(t_k))$, where $\texttt{TE}(\cdot)$ is timestep embedding following DiT~\cite{peebles2023scalable} and ADM~\cite{dhariwal2021diffusion}.
\item Historical queries $\{\mathbf{Q}_{k-i}\}_{i=1}^n$: queued past queries from the memory bank.
% \item Historical Condition $\{\mathbf{C}_{k-i}\}_{i=1}^n$: Each $\mathbf{C}_{k-i} = \texttt{MLP}([\texttt{PE}(t_{k-i}), \texttt{PE}(\Delta t_i), \texttt{PE}(\Delta k_i)])$ 
% encodes the historical noise $t_{k-i}$, relative noise scale shift $\Delta t_i = t_k - t_{k-i}$, and the frame interval $\Delta k_i$.
%\end{itemize}
\item Historical conditions $\{\mathbf{C}_{k-i}\}_{i=1}^n$: 
each condition is embedded as:
\begin{equation}
\mathbf{C}_{k-i} = \texttt{MLP}\big( [\texttt{TE}(t_{k-i}), \texttt{TE}(\Delta t_i), \texttt{TE}(\Delta k_i)] \big)
\end{equation}
where $t_{k-i}$ is the historical noise level, $\Delta t_i = t_k - t_{k-i}$ is the relative noise scale shift, and $\Delta k_i$ is the frame interval.
\end{enumerate}

\noindent{\textbf{Temporal Transition Module (TTM).}}
TTM aligns historical queries with the current denoising stage. Through an MLP, each historical condition $\mathbf{C}_{k-i}$ generates scale and shift parameters, which then transform the historical queries: 
\begin{equation}
    \begin{aligned}
        \gamma_{k-i}, \beta_{k-i} &= \texttt{MLP}(\mathbf{C}_{k-i}), \\
        \tilde{\mathbf{Q}}_{k-i} &= (1 + \gamma_{k-i}) \odot \text{LayerNorm}(\mathbf{Q}_{k-i}) + \beta_{k-i}.
    \end{aligned}
\label{eq:ttm_align}
\end{equation}
This way, the generated temporal queries $\{\tilde{\mathbf{Q}}_{k-i}\}_{i=1}^n$ effectively re-project historical features into the current noise manifold for optimized cross-frame reasoning.

\noindent{\textbf{Temporal Fusion.}}
With the inputs aligned, the interaction is executed through the Temporal Cross-Attention layer introduced in Sec.~\ref{sec:decoder}. This layer takes the modulated current queries $\tilde{\mathbf{Q}}_k$ and the TTM-aggregated features $\{\tilde{\mathbf{Q}}_{k-i}\}_{i=1}^n$ for interaction. This allows each task instance to inherit motion and structural priors from the past $n$ frames without being disrupted by the noise-level mismatch across different physical timestamps.

\subsection{Anchor Refresh Strategy}
\label{sec:ars}

% this one
\begin{algorithm}[t]
\caption{Anchor Refresh Strategy at the $j$-th Iteration}
\label{alg:ars_step}
\begin{algorithmic}[1]
\INPUT Noisy anchors $a_{t}$ at diffusion step $t$; anchor priors $\mathbf{A} = \{a^{(i)}\}_{i=1}^{N_\text{all}}$, where $N_{\text{all}} = N_{a} + N_{m}+ N_{p}$; step size $m$; confidence threshold $\tau$.
\OUTPUT Refined anchors $\{a_{t-m}^{(i)}\}_{i=1}^{N_\text{all}}$ for the subsequent diffusion step.
\STATE // \textit{Prediction and confidence estimation}
\STATE $\{\hat{y}_{t}^{(i)}, s_{t}^{(i)}\} = \Phi(\{a_{t}^{(i)}\})$;
\FOR{each anchor index $i \in \{1, \dots, N_\text{all}\}$}
    \IF{$s_{t}^{(i)} > \tau$}
        \STATE // \textit{Selective denoising for high-confidence predictions}
        \STATE $a_{t-m}^{(i)} \leftarrow \mathrm{DDIM}(a_{t}^{(i)}, \hat{y}_{t}^{(i)}, t, t-m)$;
    \ELSE
        \STATE // \textit{Anchor refresh for low-confidence predictions}
        \STATE $a_{t-m}^{(i)} \sim \mathcal{N}\!\left(\sqrt{\bar{\alpha}_{t-m}}\, a^{(i)}, \left(1 - \bar{\alpha}_{t-m}\right)\mathbf{I}\right)$;
    \ENDIF
\ENDFOR
\RETURN $\{a_{t-m}^{(i)}\}_{i=1}^{N_\text{all}}$;
\end{algorithmic}
\end{algorithm}

% \subsubsection{Sparse Supervision in Training}
% Due to the numerical disparity between the number of anchors $N_a$ and ground-truth instances $N_{gt}$, we adopt a sparse supervision strategy. We perform optimal bipartite matching $\sigma$ between the predictions and the GT set, computing the training loss exclusively for the matched pairs:
% \begin{equation}
% \mathcal{L}_{train} = \sum_{i=1}^{N_{gt}} \mathcal{L}_{task}(\Phi({a_{t}})_{\sigma(i)}, y_i), \quad {a_{t}} \in \mathbf{A}_{t}
% \end{equation} where $\Phi(\cdot)$ represents the decoder transformation. The remaining unmatched outputs are effectively ignored during backpropagation, preventing the model from being over-constrained.

\noindent\textbf{Training Inference Inconsistencies.~~}
%
%We observe that sparse query-based diffusion planning methods~\cite{liao2025diffusiondrive, zheng2025resad, yin2025diffrefiner} often suffer from a severe training-inference inconsistency: during training, only a small subset planning queries matched to the ground truth are optimized, while the remaining queries receive little to no supervision. 
%As a result, a query distribution shift emerges at inference time. 
%Specifically, although all queries are updated at each denoising step during inference, the previously unexplored queries are fed back as inputs to subsequent iterations. 
%
Sparse query-based diffusion planning approaches~\cite{liao2025diffusiondrive, yin2025diffrefiner} often suffer from severe training–inference inconsistencies. During training, only a small subset of planning queries matched to the ground truth is supervised, while the remaining queries receive little or no guidance. At inference, however, all queries are updated at each denoising step, including previously unexplored queries that are fed back into subsequent iterations.
This causes a progressive drift in the query distribution, gradually deviating from the anchor distribution observed during training. 
As identified in DiffusionDet~\cite{chen2023diffusiondet}, directly propagating these unconstrained, stochastic outputs through iterative steps causes a significant training-inference misalignment. Therefore, without explicit mechanisms to correct this misalignment, the discrepancy accumulates over denoising steps, leading to increasingly degraded predictions.
%Such behavior fundamentally contradicts the iterative refinement principle of diffusion models, where outputs are expected to progressively improve as denoising proceeds.

% Sparse supervision categorizes inference-time predictions into desired (high-fidelity, matched) and undesired (stochastic, unmatched) subsets. As identified in DiffusionDet~\cite{chen2023diffusiondet}, directly propagating these unconstrained, stochastic outputs through iterative steps causes a significant training-inference misalignment. This stems from the fact that their distributions are inherently stochastic and unconstrained, whereas the inputs during the training phase are strictly structured and constructed by adding prescribed noise to pre-defined anchors. 

\noindent\textbf{Inference via Anchor Refresh Strategy.}
Inspired by DiffusionDet~\cite{chen2023diffusiondet}, we propose an Anchor Refresh strategy (ARS) to address the training–inference inconsistencies. 
%During inference, high-confidence queries are retained, while low-confidence task queries are refreshed by resampling from the original noise distribution. 
During inference, high-confidence queries are retained, while low-confidence queries are refreshed by resampling from the original noise distribution (\cref{alg:ars_step}).
%(Algorithm~\ref{alg:ars_step}).
\begin{itemize}
    
\item {\textbf{Selective Denoising:}} At each denoising step, we identify high-confidence predictions using a task-specific threshold $\tau$. 
%During the denoising process, we identify desired predictions based on a task-specific confidence threshold $\gamma$. 
For queries with confidence $s_t^{(i)} > \tau$, we apply the standard DDIM reverse operator to transition from $a_t^{(i)}$ to $a_{t-m}^{(i)}$ using on the predicted clean state $\hat{y}_t^{(i)}$. This ensures that only task-relevant information is propagated.

\item {\textbf{Anchor-based Distribution Refresh:}} For low-confidence queries $s_t^{(i)} \le \tau$, we discard the unconstrained outputs and replace them with their initial anchors $a^{(i)}$, re-projected to the prescribed noise level at step $t-m$.
%For the positions occupied by undesired predictions ($s_t^{(i)} \le \gamma$), we implement a refresh mechanism. We discard these unconstrained outputs and replace them with their corresponding initial anchors $a^{(i)}$, re-projected back to the prescribed noise level of step $t-m$.
\end{itemize}

%By selectively retaining high-confidence predictions and refreshing uncertain ones, 
With the above operations, ARS preserves query diversity while maintaining alignment with the training distribution, enabling stable and robust inference in complex multi-modal scenarios.

%This mechanism preserves query diversity while maintaining alignment with the training distribution, ensuring stable and reliable performance even in complex multi-modal settings.

\subsection{Loss Function}
The entire UniTeD framework is trained end-to-end in a fully differentiable way. To ensure the overall performance of perception and planning in driving, we define a multi-task optimization objective that covers four primary tasks: detection, motion prediction, mapping, and planning. Each task can be optimized using both classification and regression losses with corresponding weight. The total loss $\mathcal{L}_{\text{total}}$ is defined as a weighted combination of task-specific components:
\begin{equation}\mathcal{L}_{\text{total}} = \lambda_{\text{det}} \mathcal{L}_{\text{det}} + \lambda_{\text{mot}} \mathcal{L}_{\text{mot}} + \lambda_{\text{map}} \mathcal{L}_{\text{map}} + \lambda_{\text{plan}} \mathcal{L}_{\text{plan}}.\end{equation}

\section{Experiments}

\subsection{Experimental Setup} 

\noindent{\textbf{Datasets.~~}}
We evaluate UniTeD on three benchmarks to validate mid-term simulation and interactive closed-loop performance.

\noindent\textbf{NAVSIM v1 \& v2.} Built on the nuPlan~\cite{2021nuPlan} dataset, NAVSIM~\cite{2024NAVSIM} provides a non-reactive mid-term simulation with a 4-second horizon, bridging the gap between open-loop and closed-loop evaluation.
Performance is measured via the Predictive Driver Model Score (PDMS)~\cite{2024NAVSIM} and Extended PDMS (EPDMS)~\cite{2025Pseudo}, which incorporate safety penalties %(e.g., collisions, traffic infractions) 
alongside efficiency and comfort rewards.

\noindent\textbf{Bench2Drive.} Bench2Drive~\cite{2024Bench2Drive} is a large-scale, CARLA-based~\cite{2017CARLA} closed-loop benchmark featuring 44 scenarios across 220 intersections. It tests interactive driving and long-tail robustness. We report the Driving Score (DS) and Success Rate (SR) as primary indicators.

\begin{table*}[!t]
\centering
%\caption{Comparison with state-of-the-art methods on the NAVSIMv1 benchmark. Disc, Sep, Gen and Uni denote the Discriminative, Separate, Generative and Unified. C and L denote Camera and LiDAR modalities, respectively.}
\caption{Performance comparison with other state-of-the-art methods on the NAVSIMv1 benchmark. All compared methods utilize ResNet-34 as the CNN backbone. The best results are highlighted in \textbf{bold}. The second best is \underline{underline}, and the third best is in \textit{italics}. 
Dis and Gen denote discriminative and generative formulations, respectively. Sep and Uni indicate whether perception and planning are formulated separately or in a unified manner. C and L denote Camera and LiDAR modalities.}
\label{tab:main_results}
% \fontsize{6.55}{9}\selectfont
\resizebox{0.7\textwidth}{!}  
{
\begin{tabular}{@{}l|cc|ccccc|c@{}}
    \toprule
    \multirow{2}{*}{Method} & \multirow{2}{*}{Paradigm} & \multirow{2}{*}{Modality} & \multicolumn{5}{c|}{Planning Metrics} & {Overall} \\
    \cline{4-9}
    & & & NC$\uparrow$ & DAC$\uparrow$\ & EP$\uparrow$ & TTC$\uparrow$ & COMF$\uparrow$ & PDMS$\uparrow$ \\
    \midrule
    VADv2~\cite{chen2024vadv2}  & Dis-Sep & C & 97.2 & 89.1 & 76.0 & 91.6 & 100.0 & 80.9 \\
    UniAD~\cite{hu2023planning} & Dis-Sep & C & 97.8 & 91.9 & 78.8 & 92.9 & 100.0 & 83.4 \\
    TransFuser~\cite{chitta2022transfuser} & Dis-Sep & C+L & 97.7 & 92.8 & 79.2 & 92.8 & 100.0 & 84.0 \\
    Hydra-MDP~\cite{li2024hydra} & Dis-Sep & C+L & 98.3 & 96.0 & 78.7 & 94.6 & 100.0 & 86.5 \\
    Hydra-MDP$++$~\cite{li2025hydra} & Dis-Sep & C & 97.6 & 96.0 & 80.4 & 93.1 & 100.0 & 86.6 \\
    WoTE~\cite{2025End} & Dis-Sep & C & \underline{98.5} & \textit{96.8} & 81.9 & 94.9 & 99.9 & 88.3 \\
    \midrule
    DiffusionDrive~\cite{liao2025diffusiondrive} & Gen-Sep & C+L & 98.2 & 96.2 & 82.2 & 94.7 & 100.0 & 88.1 \\
    GaussianFusion~\cite{2025GaussianFusion} & Gen-Sep & C+L & 98.3 & \underline{97.2} & \textit{83.0} & 94.6 & 100.0 & \textit{88.8} \\
    DiffRefiner~\cite{yin2025diffrefiner} & Gen-Sep & C & \textit{98.4} & \textbf{97.4} & \underline{83.4} & \textit{95.3} & 100.0 & \underline{89.4} \\
    \midrule
    PARA-Drive~\cite{weng2024drive} & Dis-Uni & C & 97.9 & 92.4 & 79.3 & 93.0 & 99.8 & 84.0 \\
    HiP-AD~\cite{jia2025drivetransformer} & Dis-Uni & C & \textbf{98.9} & 96.7 & 81.2 & \underline{96.3} & 99.9 & 88.6 \\
    \midrule
    \rowcolor{gray!15}
    \textbf{UniTeD} & Gen-Uni & C & \textbf{98.9} & \underline{97.2} & \textbf{84.1} & \textbf{96.6} & \textbf{100.0} & \textbf{90.2} \\
    \bottomrule
\end{tabular}
}
\end{table*}

\begin{table*}[!t]
\centering
%\caption{Detailed performance comparison on the NAVSIMv2 benchmark. We report extensive planning metrics including safety, comfort, and reliability. }
\caption{Performance comparison with other state-of-the-art methods on the NAVSIMv2 benchmark. The best results are highlighted in \textbf{bold}. The second best is \underline{underline}, and the third best is in \textit{italics}.}
\label{tab:detailed_comparison}
\resizebox{0.7\textwidth}{!} 
{
\begin{tabular}{@{}l|cccc|ccc|cc|c@{}}
    \toprule
    \multirow{2}{*}{Method} & \multicolumn{4}{c|}{Basic Planning $\uparrow$} & \multicolumn{3}{c|}{Safety $\uparrow$} & \multicolumn{2}{c|}{Comfort $\uparrow$} & {Overall} \\
    \cmidrule(lr){2-5} \cmidrule(lr){6-8} \cmidrule(lr){9-10} \cmidrule(l){11-11}
    & NC & DAC & DDC & TL & EP & TTC & LK & HC & EC & EPDMS$\uparrow$ \\
    \midrule
    TransFuser~\cite{chitta2022transfuser} & 96.9 & 89.9 & 97.8 & \textit{99.7} & 87.1 & 95.4 & 92.7 & \textit{98.3} & \underline{87.2} & 76.7 \\
    Hydra-MDP$++$~\cite{li2025hydra} & 97.2 & \textbf{97.5} & \underline{99.4} & 99.6 & 83.1 & 96.5 & 94.4 & 98.2 & 70.9 & 81.4 \\
    DriveSuprim~\cite{2025DriveSuprim} & 97.5 & 96.5 & \underline{99.4} & 99.6 & \textbf{88.4} & 96.6 & 95.5 & \textit{98.3} & 77.0 & 83.1 \\
    DiffusionDrive~\cite{liao2025diffusiondrive} & 98.2 & 96.2 & 98.6 & {-} & \underline{87.6} & 97.3 & \textit{97.4} & \underline{98.4} & {-} & 84.0 \\
    GaussianFusion~\cite{2025GaussianFusion} & \textit{98.3} & \textit{97.3} & \textit{99.0} & {-} & \textit{87.5} & \textit{97.4} & \textit{97.4} & \textit{98.3} & {-} & \textit{85.0} \\
    DiffRefiner~\cite{yin2025diffrefiner} & \underline{98.5} & \underline{97.4} & \textbf{99.6} & \underline{99.8} & \underline{87.6} & \underline{97.7} & \underline{97.7} & \textit{98.3} & \textit{86.2} & \underline{86.2} \\
    \midrule
    \rowcolor{gray!15}
    \textbf{UniTeD} & \textbf{99.1} & 97.2 & \textbf{99.6} & \textbf{99.9} & 86.9 & \textbf{98.5} & \textbf{98.4} & \textbf{99.9} & \textbf{87.3} & \textbf{90.1} \\
    \bottomrule
\end{tabular}
}
\end{table*}

% \subsection{Datasets and Metrics} We evaluate UniTeD across three representative benchmarks to validate its performance in both mid-term simulation and interactive closed-loop scenarios.
% \paragraph{NAVSIMv1.} Built on the nuPlan~\cite{2021nuPlan} dataset, NAVSIMv1~\cite{2024NAVSIM} provides a non-reactive mid-term simulation with a 4-second horizon, bridging the gap between open-loop and closed-loop evaluation. We adopt the Predictive Driver Model (PDM) score as the primary metric to assess planning quality beyond simple displacement errors.
% \paragraph{NAVSIMv2.} As an evolution for the 2025 challenge, NAVSIMv2~\cite{2025Pseudo} incorporates a pseudo closed-loop aggregation mechanism. We utilize the Extended PDM Score (EPDMS) as the core metric, which integrates safety criticalities, such as collisions and traffic light infractions, as multiplicative penalties while rewarding efficiency and comfort.
% \paragraph{Bench2Drive.} Bench2Drive~\cite{2024Bench2Drive} is a large-scale CARLA-based closed-loop benchmark~\cite{2017CARLA} featuring 44 scenarios across 220 intersections. It evaluates the planner's ability to handle interactive environments and long-tail cases. We report the Driving Score (DS) and Success Rate (SR) as key performance indicators.

\noindent\textbf{Implementation Details.~~} 
We employ ResNet-34 as the backbone for feature extraction. Input images are resized to a resolution of $640 \times 352$. For the NAVSIM benchmark~\cite{2024NAVSIM,2025Pseudo}, the model processes inputs from eight multi-view cameras, while for Bench2Drive~\cite{2024Bench2Drive}, six multi-view cameras are utilized.
Our decoder consists of six cascaded diffusion decoder layers. Following the design of DiffusionDrive~\cite{liao2025diffusiondrive}, we adopt a truncated diffusion policy operating on 1,024 object anchors, categorized into 900 agent anchors, 100 map anchors, and 24 planning anchors.
During training, the diffusion schedule is truncated at 50 out of 1,000 steps to diffuse the anchors. For inference, we use 2 denoising steps with step size $m$ of 10 and select the top-1 scoring trajectory as the final prediction. The confidence threshold $\tau$ for each task is detailed in the Appendix.
The model is trained on 8 NVIDIA L40 GPUs with a total batch size of 64. We use the AdamW optimizer with an initial learning rate of $4 \times 10^{-4}$ and train for 100 epochs to ensure convergence across benchmarks.

\subsection{Main Results}

% \begin{table}[tb] % 改回单栏环境，更紧凑
% \centering
% %\caption{Comparison on the Bench2Drive online leaderboard. We report the Driving Score (DS) and Success Rate (SR). }
% \caption{Comparison on the Bench2Drive online leaderboard.}
%   \label{tab:bench2drive_results}
%   % \small 
%   % \renewcommand{\arraystretch}{1.1} % 稍微收紧行高
%   % \setlength{\tabcolsep}{8pt} % 缩短列间距，使其符合单栏宽度
%   \resizebox{0.67\textwidth}{!} 
%   {
%   \begin{tabular}{@{}llcc} % 彻底去掉竖线，完美解决 rowcolor 溢出
%     \toprule
%     Method & Training Dataset & DS $\uparrow$ & SR (\%) $\uparrow$ \\
%     \midrule
%     VAD~\cite{jiang2023vad}  & B2D (200K) & 42.2 & 15.0 \\
%     UniAD~\cite{hu2023planning} & B2D (200K) & 45.8 & 16.4 \\
%     SparseDrive~\cite{sun2025sparsedrive} & B2D (200K) & 44.5 & 16.7 \\
%     DriveTransformer~\cite{jia2025drivetransformer} & B2D (200K) & 63.5 & 35.0 \\
%     HiP-AD~\cite{tang2025hip} & B2D (200K) & 86.8 & 69.1 \\
%     \midrule
%     TF++~\cite{Zimmerlin2024HiddenBO} & TF++ (500K) & 84.2 & 67.3 \\
%     DiffusionDrive~\cite{liao2025diffusiondrive} & TF++ (500K) & 77.7 & 52.7 \\
%     DiffRefiner~\cite{yin2025diffrefiner} & TF++ (500K) & 87.1 & \textbf{71.4} \\
%     \midrule
%     \rowcolor{gray!15}
%     \textbf{UniTeD} & B2D (200K) & \textbf{87.3} & 70.0 \\
%     \bottomrule
%   \end{tabular}
%   }
% \end{table}

\noindent{\textbf{Results on NAVSIM v1.}} To ensure a fair comparison, all baseline models are evaluated using a ResNet-34 backbone.
Methods based on reinforcement learning or Vision-Language Models (VLMs) are excluded from the main analysis, with detailed comparisons provided in the Appendix. 
As summarized in~\cref{tab:main_results}, our proposed UniTeD achieves state-of-the-art performance on the NAVSIM v1 benchmark, yielding a PDMS of 90.2. 
%while methods based on reinforcement learning or Vision-Language Models (VLMs) are excluded from our analysis, with detailed comparisons provided in the Appendix. 
UniTeD consistently outperforms existing approaches across different methodological paradigms. Specifically, UniTeD surpasses the best discriminative-separate models, including WoTE~\cite{2025End} (88.3) and Hydra-MDP++~\cite{li2025hydra} (86.6), by margins of +1.9 and +3.6 PDMS, respectively. These improvements highlight the advantage of generative modeling over deterministic regression in capturing complex and diverse driving behaviors. Compared with the diffusion-based DiffRefiner~\cite{yin2025diffrefiner} (89.4), which adopts a generative-separate pipeline, UniTeD still achieves a further gain of +0.8 PDMS. 
In comparison to the unified discriminative counterpart HiP-AD~\cite{tang2025hip} (88.6), our method delivers a +1.6 improvement, demonstrating the effectiveness of formulating both perception and planning queries within a unified diffusion framework.
Notably, relying solely on camera (C) inputs, UniTeD outperforms all competitive LiDAR-based (C+L) approaches, demonstrating its efficacy and robustness.

%including GaussianFusion~\cite{2025GaussianFusion} (88.8) and DiffusionDrive~\cite{liao2025diffusiondrive} (88.1), by +1.4 and +2.1 PDMS, respectively. This result further demonstrates the robustness and efficiency of our framework.
%Remarkably, while relying solely on Camera (C) inputs, UniTeD surpasses all competitive LiDAR-based (C+L) methods, including GaussianFusion~\cite{2025GaussianFusion} (88.8) and DiffusionDrive~\cite{liao2025diffusiondrive} (88.1) by +1.4 and +2.1 points. This demonstrates the efficacy and robustness of our framework.   

\noindent{\textbf{Results on NAVSIM v2.}} %As presented in ~\cref{tab:detailed_comparison}, UniTeD maintains its superior performance under the more stringent NAVSIM v2 benchmark, achieving a state-of-the-art EPDMS of 90.1. Specifically, our method achieves a remarkable lead of +3.9 points over the strongest generative competitor, DiffRefiner~\cite{yin2025diffrefiner} (86.2). These consistent and significant improvements across all key competitors demonstrate the superior planning capability and robustness of our proposed framework in handling complex, real-world driving scenarios.
As shown in~\cref{tab:detailed_comparison}, UniTeD maintains its superior performance under the more challenging NAVSIM v2 benchmark, achieving a state-of-the-art EPDMS of 90.1. In particular, our method outperforms the strongest generative baseline, DiffRefiner~\cite{yin2025diffrefiner} (86.2), by a substantial margin of +3.9 points. Such consistent and significant gains over all major competitors further validate the enhanced planning capability and robustness of our framework in complex, real-world driving scenarios.

\noindent{\textbf{Results on Bench2Drive.}}
To evaluate the closed-loop planning capability of our model in reactive environments, we conduct experiments on the Bench2Drive leaderboard. As reported in~\cref{tab:bench2drive_results}, UniTeD achieves a state-of-the-art Driving Score (DS) of 87.3, outperforming all existing approaches.
When trained on the standard Bench2Drive~\cite{2024Bench2Drive} 200K dataset, UniTeD consistently surpasses contemporary unified methods, including HiP-AD~\cite{tang2025hip} (86.8 DS) and DriveTransformer~\cite{jia2025drivetransformer} (63.5 DS), achieving improvements of +0.5 and +23.8 DS, respectively. Notably, even compared with models trained on the substantially larger and higher-quality TF++~\cite{Zimmerlin2024HiddenBO} (500K) dataset, UniTeD maintains its advantage. It outperforms the recent generative model DiffRefiner~\cite{yin2025diffrefiner} by +0.2 DS and DiffusionDrive~\cite{liao2025diffusiondrive} by +9.6 DS.
These results highlight the strong robustness and generalization capability of UniTeD. 
%Despite being trained with 60\% less data than TF++-based methods, our diffusion-based planner demonstrates superior adaptability to complex and dynamic traffic scenarios.

% \begin{table}[tb]
% \centering
% %\caption{Perception and Prediction performance on the nuScenes validation set. $\uparrow$ and $\downarrow$ denote higher and lower values are better, respectively.}
% \caption{Perception and Prediction performance on the nuScenes validation set.}
% \label{tab:nuscenes_perception_refined}
% % 使用与模板一致的缩放方式
% \resizebox{0.72\textwidth}{!} 
% {
% \begin{tabular}{@{}l|cc|c|c|c@{}}
%     \toprule
%     \multirow{2}{*}{Method} & \multicolumn{2}{c|}{Detection} & {Map} & {Track} & {Motion} \\
%     \cline{2-6}
%     & mAP$\uparrow$ & NDS$\uparrow$ & mAP$\uparrow$ & AMOTA$\uparrow$ & minADE$\downarrow$ \\
%     \midrule
%     UniAD~\cite{hu2023planning}   & 0.380 & 0.359 & - & - & - \\
%     VAD~\cite{jiang2023vad}      & 0.276 & 0.397 & 0.476 & - & - \\
%     SparseDrive-S~\cite{sun2025sparsedrive} & 0.418 & 0.525 & 0.551 & 0.386 & 0.62 \\
%     DiFSD~\cite{2024DiFSD}        & 0.410 & 0.528 & 0.560 & - & - \\
%     HiP-AD~\cite{tang2025hip}      & \textbf{0.424} & 0.535 & 0.571 & 0.406 & 0.61 \\
%     \midrule
%     \rowcolor{gray!15}
%     \textbf{UniTeD} & \textbf{0.424} & \textbf{0.537} & \textbf{0.596} & \textbf{0.419} & \textbf{0.58} \\
%     \bottomrule
% \end{tabular}
% }
% \end{table}

\begin{table*}[t] % 如果是双栏文档请用 table*，单栏用 table
  \centering
  
  % --- 第一个表格 (Bench2Drive) ---
  \begin{minipage}[b]{0.45\textwidth}
    \centering
    \caption{Comparison on the Bench2Drive online leaderboard.}
    \label{tab:bench2drive_results}
    \resizebox{\textwidth}{!}{
      \begin{tabular}{@{}llcc}
        \toprule
        Method & Training Data  & DS $\uparrow$ & SR(\%) $\uparrow$ \\
        \midrule
        VAD~\cite{jiang2023vad}  & B2D (200K) & 42.2 & 15.0 \\
        UniAD~\cite{hu2023planning} & B2D (200K) & 45.8 & 16.4 \\
        SparseDrive~\cite{sun2025sparsedrive} & B2D (200K) & 44.5 & 16.7 \\
        DriveTransformer~\cite{jia2025drivetransformer} & B2D (200K) & 63.5 & 35.0 \\
        HiP-AD~\cite{tang2025hip} & B2D (200K) & 86.8 & 69.1 \\
        \midrule
        TF++~\cite{Zimmerlin2024HiddenBO} & TF++ (500K) & 84.2 & 67.3 \\
        DiffusionDrive~\cite{liao2025diffusiondrive} & TF++ (500K) & 77.7 & 52.7 \\
        DiffRefiner~\cite{yin2025diffrefiner} & TF++ (500K) & 87.1 & \textbf{71.4} \\
        \midrule
        \rowcolor{gray!15}
        \textbf{UniTeD} & B2D (200K) & \textbf{87.3} & 70.0 \\
        \bottomrule
      \end{tabular}
    }
  \end{minipage}
  \hfill % 自动分配中间间距
  % --- 第二个表格 (nuScenes) ---
  \begin{minipage}[b]{0.5\textwidth}
    \centering
    \caption{Perception and Prediction performance on the nuScenes validation set.}
    \label{tab:nuscenes_perception_refined}
    \resizebox{\textwidth}{!}{
      \begin{tabular}{@{}l|cc|c|c|c@{}}
        \toprule
        \multirow{2}{*}{Method} & \multicolumn{2}{c|}{Detection} & {Map} & {Track} & {Motion} \\
        \cline{2-6}
        & mAP$\uparrow$ & NDS$\uparrow$ & mAP$\uparrow$ & AMOTA$\uparrow$ & minADE$\downarrow$ \\
        \midrule
        UniAD~\cite{hu2023planning}   & 0.380 & 0.359 & - & - & - \\
        VAD~\cite{jiang2023vad}      & 0.276 & 0.397 & 0.476 & - & - \\
        SparseDrive-S~\cite{sun2025sparsedrive} & 0.418 & 0.525 & 0.551 & 0.386 & 0.62 \\
        DiFSD~\cite{2024DiFSD}        & 0.410 & 0.528 & 0.560 & - & - \\
        HiP-AD~\cite{tang2025hip}      & \textbf{0.424} & 0.535 & 0.571 & 0.406 & 0.61 \\
        \midrule
        \rowcolor{gray!15}
        \textbf{UniTeD} & \textbf{0.424} & \textbf{0.537} & \textbf{0.596} & \textbf{0.419} & \textbf{0.58} \\
        \bottomrule
      \end{tabular}
    }
  \end{minipage}

\end{table*}

\noindent{\textbf{Results on nuScenes.}}
%Beyond planning performance, we further evaluate the perception capability of UniTeD on the nuScenes~\cite{2019nuScenes} open-loop benchmark. 
We further evaluate the perception capability of UniTeD on nuScenes~\cite{2019nuScenes}. 
As shown in~\cref{tab:nuscenes_perception_refined}, the results demonstrate that UniTeD is not only a strong planner but also a versatile framework capable of maintaining high-fidelity environmental understanding.
Specifically, UniTeD achieves 0.596 mAP on map segmentation, outperforming HiP-AD~\cite{tang2025hip} by +2.5\% in map mAP. 
%This improvement suggests that our unified generative representation captures more precise spatial semantics and structural constraints of the environment. 
For 3D object detection, it attains a leading 0.537 NDS. In motion prediction, UniTeD achieves the lowest minADE of 0.58. Compared with SparseDrive-S~\cite{sun2025sparsedrive}, it reduces motion prediction error by 6.4\% (from 0.62 to 0.58).
Moreover, UniTeD reaches 0.419 AMOTA, surpassing HiP-AD by +1.3\% and SparseDrive-S by +3.3\%. These results collectively confirm that UniTeD serves as a comprehensive and unified framework for perception, prediction, and planning.

\begin{table}[tb]
%\caption{Ablation study of different paradigms and task components.Sep., Regr., Uni. and Diff. denote the Separate, Regression, Unified and Diffusion.}
\caption{
Ablation study of different paradigms and task components. 
Perc and Plan denote perception and planning modules.
Sep and Uni indicate whether perception and planning are formulated separately or in a unified manner.
Regr and Diff denote regression-based and diffusion-based approaches, respectively.
}
  \label{tab:headings}
  \centering
  \small 
  \renewcommand{\arraystretch}{1.3} 
  \setlength{\tabcolsep}{3pt} 
  \resizebox{0.57\textwidth}{!} 
{
  \begin{tabular}{c | cc | cc | S[table-format=2.1] S[table-format=2.1] S[table-format=2.1] S[table-format=2.1] S[table-format=3.1] S[table-format=2.1]}
    \toprule
    \multirow{2}{*}{ID} & \multicolumn{2}{c|}{Paradigm} & \multirow{2}{*}{Perc} & \multirow{2}{*}{Plan} & \multicolumn{6}{c}{NAVSIMv1 Metrics} \\
    \cline{2-3} \cline{6-11}
    & Sep & Uni & & & {NC} & {DAC} & {EP} & {TTC} & {C} & {PDMS} \\
    \midrule
    0 & \cmark & & Regr & Regr & 97.8 & 91.9 & 79.2 & 92.8 & 100.0 & 84.1 \\
    1 & \cmark & & Regr & Diff & 97.6 & 96.0 & 80.4 & 93.1 & 100.0 & 86.7 \\
    2 & & \cmark & Regr & Regr & 98.8 & 96.7 & 81.1 & 96.3 & 99.9 & 88.5 \\
    3 & & \cmark & Regr & Diff & 98.8 & 97.0 & 82.7 & \textbf{96.7} & 99.9 & 89.5 \\
    4 & & \cmark & Diff & Diff & \textbf{99.0} & \textbf{97.2} & \textbf{84.1} & 96.6 & \textbf{100.0} & \textbf{90.2} \\
    \bottomrule
  \end{tabular}
}
\end{table}

\begin{table}[tb]
%\caption{Ablation study of TTM. Regr., Uni., Diff., Memory and TTM denote the Regression, Unified, Diffusion, Memory Queue and Temporal Transition Module.}
\caption{Ablation study of TTM in unified paradigm. 
Regr and Diff are defined the same as in~\cref{tab:headings}.
Memory denotes the Memory Queue.}
  \label{tab:ablation_study}
  \centering
  \small 
  \renewcommand{\arraystretch}{1.2} 
  \setlength{\tabcolsep}{3.8pt} % 稍微增加间距，使视觉不拥挤
  \resizebox{0.7\textwidth}{!} 
{
  \begin{tabular}{c | cc | cc | S[table-format=2.1] S[table-format=2.1] S[table-format=2.1] S[table-format=2.1] S[table-format=3.1] S[table-format=2.1]}
    \toprule
    \multirow{2}{*}{ID} & \multicolumn{2}{c|}{Unified Paradigm} & \multirow{2}{*}{Memory} & \multirow{2}{*}{TTM} & \multicolumn{6}{c}{NAVSIMv1 Metrics} \\
    \cline{2-3} \cline{6-11}
    & Regr & Diff & & & {NC} & {DAC} & {EP} & {TTC} & {C} & {PDMS} \\
    \midrule
    0 & \cmark &        & \xmark & {-}  & 98.9 & 96.0 & 81.0 & 96.4 & 99.9 & 88.2 \\
    1 & \cmark &        & \cmark & {-}  & 98.8 & 96.7 & 81.1 & 96.3 & 99.9 & 88.5 \\
    \midrule
    2 &        & \cmark & \xmark & \xmark & 98.8 & 96.8 & 82.2 & 96.4 & 99.9 & 89.0 \\
    3 &        & \cmark & \cmark & \xmark & 99.0 & 97.1 & 82.5 & 96.5 & 99.9 & 89.4 \\
    4 &        & \cmark & \cmark & \cmark & \textbf{99.0} & \textbf{97.2} & \textbf{84.1} & \textbf{96.6} & \textbf{100.0} & \textbf{90.2} \\
    \bottomrule
  \end{tabular}
}
\end{table}

\begin{table}[!t]

  \caption{Ablation study on the Anchor Refresh Strategy.}
  \label{tab:ablation_refresh}
  \centering
  \small 
  \renewcommand{\arraystretch}{1.2} 
  \setlength{\tabcolsep}{8pt} 
  \resizebox{0.7\textwidth}{!} 
{
  \begin{tabular}{c | c | S[table-format=2.1] S[table-format=2.1] S[table-format=2.1] S[table-format=2.1] S[table-format=3.1] S[table-format=2.1]}
    \toprule
    \multirow{2}{*}{ID} & \multirow{2}{*}{ARS} & \multicolumn{6}{c}{NAVSIMv1 Metrics} \\
    \cline{3-8}
    & & {NC } & {DAC } & {EP } & {TTC} & {C} & {PDMS} \\
    \midrule
    0 & \xmark & 98.4 & 96.8 & 81.7 & 95.2 & 99.9 & 88.2 \\
    1 & \cmark & \textbf{99.0} & \textbf{97.2} & \textbf{84.1} & \textbf{96.6} & \textbf{100.0} & \textbf{90.2} \\
    \bottomrule
  \end{tabular}
}
\end{table}

\subsection{Ablation Study}

\noindent{\textbf{Effect of Unified Diffusion.~}}
We conduct a comprehensive ablation study to validate the effectiveness of each design component, as reported in~\cref{tab:headings}. By progressively incorporating the unified architecture and diffusion-based modeling, performance steadily improves, ultimately reaching a peak PDMS of 90.2.
Specifically, using the diffusion-based planning module, comparing the separate perception–planning setup to the unified perception–planning framework (ID 1 vs. ID 3) shows that transitioning to a unified paradigm yields a substantial gain of +2.8 PDMS.
This result confirms that joint learning across perception and planning alleviates the information bottleneck and mitigates error accumulation commonly observed in sequential pipelines. Under the separate paradigm (ID 0 vs. ID 1), introducing diffusion-based planning improves PDMS from 84.1 to 86.7 (+2.6). Similarly, within the unified paradigm (ID 2 vs. ID 3), diffusion modeling provides an additional +1.0 gain. These findings highlight the advantage of diffusion processes over deterministic regression in modeling the inherently multi-modal nature of human driving behavior.
Finally, extending diffusion modeling to both perception and planning (ID 4) achieves the best overall performance, surpassing the variant with diffusion-based planning only (ID 3) by +0.7 PDMS. This demonstrates that diffusion-based perception yields more robust and high-fidelity feature representations, which in turn establish a stronger foundation for downstream planning. Overall, the synergy introduced by the Unified Diffusion framework enables the model to achieve both precise environmental understanding and safe, human-like trajectory generation, outperforming conventional regression-based unified approaches.

\noindent{\textbf{Effect of TTM.~}}
We systematically analyze the components of TTM in~\cref{tab:ablation_study}. The results confirm that both temporal context modeling and timestep alignment are critical for robust planning.
Under the regression paradigm (ID 0 vs. ID 1), incorporating Memory Queue yields a +0.3 PDMS improvement. A consistent gain is also observed in the diffusion paradigm (ID 2 vs. ID 3), where PDMS increases from 89.0 to 89.4 (+0.4). The results show that historical information provides essential temporal continuity for both perception and planning, regardless of the underlying framework. Notably, even without temporal memory, the diffusion-based model (ID 2, 89.0) already outperforms the memory-enhanced regression counterpart (ID 1, 88.5) by +0.5 PDMS, further highlighting the superior capacity of generative modeling in capturing complex driving distributions.
The most significant improvement arises from the full TTM design. Comparing ID 3 and ID 4, introducing TTM increases PDMS from 89.4 to 90.2. 
%a significant gain of +0.8. 
While the Memory Queue supplies raw historical context, TTM addresses feature–denoising misalignment by synchronizing latent representations with the current diffusion timestep $t$, thereby enabling more coherent temporal reasoning.

\noindent{\textbf{Effect of Anchor Refresh Strategy.~}}
We examine the impact of ARS in~\cref{tab:ablation_refresh}. The results show that ARS is crucial for maintaining consistency within the sparse diffusion architecture. Without this mechanism (ID 0), the model relies solely on the initial trunk anchors, leading to a noticeable training–inference discrepancy. 
%As shown in Table 7, comparing ID 0 and ID 1, introducing Anchor Refresh improves PDMS from 88.2 to 90.2 (+2.0).
%These findings indicate that Anchor Refresh is not merely an auxiliary optimization, but a fundamental component for ensuring stable and consistent input distribution in UniTeD.
Introducing ARS improves PDMS from 88.2 to 90.2 (ID 0 vs. ID 1).
These findings indicate that ARS is a fundamental component for ensuring stable and consistent input distribution.

\section{Conclusion}

In this paper, we introduced UniTeD, a unified diffusion framework that seamlessly integrates perception and planning within a shared generative denoising process. By enabling bidirectional interaction and mutual refinement, UniTeD alleviates error propagation and strengthens multi-modal reasoning.
We further extend the framework to a streaming setting through a Temporal Transition Module that resolves inter-frame noise mismatch, together with an Anchor Refresh Strategy to mitigate training–inference distribution shift. Extensive experiments on challenging benchmarks demonstrate that UniTeD achieves state-of-the-art performance, surpassing both discriminative end-to-end and planning-only diffusion approaches.

\bibliographystyle{splncs04}
\bibliography{main}

%\clearpage
\appendix
\setcounter{table}{0}   
\setcounter{figure}{0}
\renewcommand{\thetable}{S\arabic{table}}
\renewcommand{\thefigure}{S\arabic{figure}}
\renewcommand{\INPUT}{\item[\textbf{Input:}] }
\renewcommand{\OUTPUT}{\item[\textbf{Output:}] }
\renewcommand{\algorithmicrequire}{\textbf{Input:}}
\renewcommand{\algorithmicensure}{\textbf{Output:}}

\title{Appendix}

\titlerunning{UniTeD for Autonomous Driving}
\author{ }
\authorrunning{B.~Zhao et al.}
\institute{ }

\maketitle
This supplementary material is the Appendix referenced in the main manuscript. It includes additional implementation details, comprehensive experimental comparisons, and visualization analysis to support the main findings. In~\cref{sec:implementation}, we describe the implementation details, covering datasets, evaluation metrics, and training and inference configurations.
Subsequently, in~\cref{sec:more-exp}, we present further experimental results and extend the comparison to additional methods. 
Finally, more visual results and corresponding analyses are provided in~\cref{sec:vis}.

%======================================
\section{More Implementation Details}
\label{sec:implementation}
\subsection{Datasets and Metrics}
\label{app:datasets_metrics}
To evaluate the performance of UniTeD in both forecasting and closed-loop interactive scenarios, we conduct experiments on three representative benchmarks: NAVSIM v1~\cite{2024NAVSIM}, NAVSIM v2~\cite{cao2025pseudo}, and Bench2Drive~\cite{2024Bench2Drive}.

\subsubsection{NAVSIM}
NAVSIM~\cite{2024NAVSIM} is a large-scale benchmark specifically designed for end-to-end (E2E) autonomous driving. It provides a diverse range of real-world driving logs, enabling a rigorous evaluation of model under complex urban scenarios.

\noindent \textbf{PDM Score (PDMS).} The official NAVSIM v1 benchmark comprises five sub-metrics: No-at-fault Collision (NC), Drivable Area Compliance (DAC), Time-to-Collision (TTC), Ego Vehicle Progress (EP), and History Comfort (C, referred to as HC in NAVSIM v2). The overall Predictive Driver Model Score (PDMS) is computed as:
\begin{equation}
\text{PDMS} = \text{NC} \times \text{DAC} \times \frac{5 \times \text{TTC} + 2 \times \text{C} + 5 \times \text{EP}}{12}.
\end{equation}
\noindent \textbf{Extended PDM Score (EPDMS).} The NAVSIM v2 benchmark \cite{cao2025pseudo} extends the evaluation protocol by introducing additional sub-metrics, including Driving Direction Compliance (DDC), Traffic Light Compliance (TC), Lane Keeping (LK), and Extended Comfort (EC), combined with a False-Positive Penalty Filtering mechanism. The final Extended Predictive Driver Model Score (EPDMS) is computed as follows:
\begin{equation}
\text{P} = \text{NC} \times \text{DAC} \times \text{DDC} \times \text{TC}
,\end{equation}
\begin{equation}
\bar{M} = \frac{\sum_{m \in \{\text{TTC}, \text{EP}, \text{HC}, \text{LK}, \text{EC}\}} w_m\times m}{\sum_{m \in \{\text{TTC}, \text{EP}, \text{HC}, \text{LK}, \text{EC}\}} w_m},
\end{equation}
\begin{equation}
\text{EPDMS} = \text{P} \times \bar{M},
\end{equation}
where P denotes the product of core compliance metrics, and $\bar{M}$ represents the weighted average of supplementary driving behavior indicators. Following the standard protocol, the weights are set as $w_{\text{TTC}} = 5, w_{\text{EP}} = 5, w_{\text{HC}} = 2, w_{\text{LK}} = 2,$ and $w_{\text{EC}} = 2$.

\subsubsection{Bench2Drive}
Bench2Drive~\cite{2024Bench2Drive} is a large-scale, CARLA-based closed-loop benchmark featuring 44 scenarios across 220 intersections.

\noindent \textbf{Success Rate (SR).} The fraction of routes completed without any terminal infractions (e.g., collisions or driving off-road):
\begin{equation}
\text{SR} = \frac{1}{N} \sum_{i=1}^{N} S_i,
\end{equation}
where N is the total number of test routes. $S_i$ is an indicator variable that equals $1$ if the $i$-th route is completed successfully, and $0$ otherwise.

\noindent \textbf{Driving Score (DS).} The primary ranking metric in Bench2Drive, defined as the average product of Route Completion ($R_i$) and Infraction Penalty ($P_i$):
\begin{equation}
    \text{DS} = \frac{1}{N} \sum_{i=1}^N (R_i \times P_i).
\end{equation}
The infraction penalty $P_i$ is a cumulative product of multipliers $p_{i,j}$ for all violations recorded in scenario $i$: $P_i = \prod_{j} p_{i,j}$. Following the official definition, the penalty coefficients $p_{i,j}$ are:
\begin{itemize}
    \item Pedestrian Collision: 0.50
    \item Vehicle Collision: 0.60
    \item Static Obstacle Collision: 0.65
    \item Stop Sign Infraction/Running Red Light/Scenario Timeout/Too Slow/No Give Way: 0.70
\end{itemize}

\subsection{Training Objective and Loss Functions}

The UniTeD framework is optimized using a multi-task learning objective. The total loss $\mathcal{L}_{\text{total}}$ is a weighted combination of four task-specific components: detection, mapping, motion prediction, and planning. Each task comprises a classification loss ($\mathcal{L}_{\text{cls}}$) and a regression loss ($\mathcal{L}_{\text{reg}}$):
\begin{equation}\mathcal{L}_{\text{total}} = \lambda_{\text{det}} \mathcal{L}_{\text{det}} + \lambda_{\text{mot}} \mathcal{L}_{\text{mot}} + \lambda_{\text{map}} \mathcal{L}_{\text{map}} + \lambda_{\text{plan}} \mathcal{L}_{\text{plan}}.\end{equation}
The detailed formulations for each component are described below:
\begin{itemize}

%\item Detection Loss ($\mathcal{L}_{\text{det}}$): We employ a Focal Loss ($\gamma=2.0, \alpha=0.25$) for object classification with a weight of 2.0. The regression branch uses a SparseBox3DLoss, which integrates an L1 Loss for bounding box coordinates (weight 0.25), a Cross-Entropy Loss for centerness estimation, and a Gaussian Focal Loss for orientation (yawness) supervision.

\item Detection Loss ($\mathcal{L}_{\text{det}}$): Object classification is supervised using focal loss ($\gamma=2.0$, $\alpha=0.25$) with a weight of 2.0. The regression branch employs a SparseBox3DLoss~\cite{lin2023sparse4d}, which combines $\ell_1$ loss for bounding box coordinates (weight 0.25), Cross-entropy loss for centerness estimation, and Gaussian focal loss for orientation (yaw) supervision.

\item Mapping Loss ($\mathcal{L}_{\text{map}}$): For online map elements, we utilize a focal loss for classification (weight 1.0). To supervise the geometric structure, an $\ell_1$ loss is applied to the sampled points along the polylines. We assign a high regression weight of 10.0 to these coordinates to ensure the precise generation of lanes and boundaries.
\item Motion Prediction Loss ($\mathcal{L}_{\text{mot}}$): The motion of surrounding agents is supervised via a focal loss for intent classification and an $\ell_1$ loss for trajectory waypoint regression. Both components are assigned a weight of 0.2 to balance their contribution relative to the ego-planning task.
\item Planning Loss ($\mathcal{L}_{\text{plan}}$): The ego-planning consists of a focal Loss for trajectory selection (weight 0.5) and an $\ell_1$ loss for the refinement of the future waypoints (weight 1.0). This combination ensures that the generated trajectories are both semantically reasonable and geometrically accurate.
\end{itemize}
All tasks are trained end-to-end, with the loss weights $\lambda$ empirically balanced to ensure the stable convergence of both the perception and the planning.

\subsection{Inference Details}
UniTeD utilizes an accelerated reverse diffusion process to achieve real-time performance.
\begin{itemize}
\item Step Schedule: During inference, the total diffusion step is set to $T=20$ steps. %We employ an accelerated 2-step DDIM sampler (with a step size of $m=10$) to ensure real-time performance.
We utilize an accelerated 2-step DDIM sampler with a step size of $m=10$ to enable real-time performance.
\item Truncated Initialization: Following the design of DiffusionDrive~\cite{liao2025diffusiondrive}, the denoising starts by applying 8 steps of diffusion noise to the anchor priors $\mathbf{A}$. This initialization at  $t=8$ provides a structural prior that significantly accelerates convergence compared to pure Gaussian noise.
\item Task-Specific Thresholds: To balance recall and safety, the Anchor Refresh Strategy (ARS) applies heterogeneous confidence thresholds $\tau$ in different tasks.
Specifically, we set $\tau = 0.30$ for detection, $\tau = 0.45$ for mapping, and $\tau = 0.30$ for both motion prediction and planning tasks.
 
\end{itemize}
% \begin{equation}
% \tau =
% \begin{cases}
% 0.30, & \text{Detection } \\
% 0.45, & \text{Mapping } \\
% 0.50, & \text{Motion \& Planning }
% \end{cases}
% \end{equation}
%======================================
\section{More Experimental Results}
\label{sec:more-exp}
\begin{table*}[!t]
\centering
\caption{Comparison with VLMs-based and RL-based methods on the NAVSIM v1 benchmark. The best results are highlighted in \textbf{bold}, and the second best are \underline{underline}.}
\label{tab:appendix_comparison}
\resizebox{0.7\textwidth}{!}{
\begin{tabular}{@{}l|ccccc|c@{}}
    \toprule
    \multirow{2}{*}{Method} & \multicolumn{5}{c|}{Planning Metrics} & Overall \\
    \cline{2-7}
     & NC$\uparrow$ & DAC$\uparrow$ & EP$\uparrow$ & TTC$\uparrow$ & COMF$\uparrow$ & PDMS$\uparrow$ \\
    \midrule
    \multicolumn{7}{l}{\textbf{VLMs-based Methods (SFT)}} \\
    Qwen2.5-VL~\cite{bai2025qwen2}    & 97.8 & 92.1 & 78.3 & 92.8 & \textbf{100.0} & 83.3 \\
    InternVL3~\cite{zhu2025internvl3} & 97.0 & 92.4 & 78.8 & 91.8 & \textbf{100.0} & 83.3 \\
    ImagiDrive~\cite{li2025imagidrive}    & 98.1 & 96.2 & 80.1 & 94.4 & \textbf{100.0} & 86.9 \\
    AutoVLA~\cite{zhou2025autovla}        & 96.9 & 92.4 & 75.8 & 88.1 & 99.1 & 80.5 \\
    Recogdrive~\cite{li2025recogdrive}    & 98.1 & 94.7 & 80.9 & 94.2 & \textbf{100.0} & 86.5 \\
    FLARE~\cite{zhang2025future}        & 98.2 & 95.0 & 81.8 & 93.9 & \textbf{100.0} & 86.9 \\
    SGDrive~\cite{li2026sgdrive}        & \underline{98.6} & 95.1 & 81.2 & 95.4 & \textbf{100.0} & 87.4 \\
    \midrule
    \multicolumn{7}{l}{\textbf{VLMs-based Methods (RFT)}} \\
    AdaThinkDrive~\cite{luo2025adathinkdrive} & 98.4 & 97.8 & 84.4 & 95.2 & \textbf{100.0} & 90.3 \\
    AutoVLA~\cite{zhou2025autovla}        & 98.4 & 95.6 & 81.9 & \textbf{98.0} & \underline{99.9} & 89.1 \\
    Recogdrive~\cite{li2025recogdrive}    & 97.9 & 97.3 & \underline{87.3} & 94.9 & \textbf{100.0} & 90.8 \\
    FLARE~\cite{zhang2025future}        & 98.5 & \textbf{98.4} & 86.0 & 96.0 & \textbf{100.0} & \textbf{91.4} \\
    SGDrive~\cite{li2026sgdrive}        & \underline{98.6} & 97.8 & 85.8 & 96.2 & \textbf{100.0} & 91.1 \\
    \midrule
    \multicolumn{7}{l}{\textbf{RL-based Methods}} \\
    Trajhf~\cite{li2025learning}            & 96.6 & 96.6 & 84.5 & 92.1 & \textbf{100.0} & 87.6 \\
    DriveDPO~\cite{shang2025drivedpo}         & 98.5 & \underline{98.1} & 84.3 & 94.8 & \underline{99.9} & 90.0 \\
    DiffusionDriveV2~\cite{zou2025diffusiondrivev2}         & 98.3 & 97.9 & \textbf{87.5} & 94.8 & \underline{99.9} & \underline{91.2} \\
    \midrule
    \rowcolor{gray!15}
    \textbf{UniTeD} & \textbf{98.9} & 97.2 & 84.1 & \underline{96.6} & \textbf{100.0} & 90.2 \\
    \bottomrule
\end{tabular}
}
\end{table*}

\subsection{Further Comparisons with SOTA Methods}
\label{app:further_comparisons}
%As shown in Table~\ref{tab:appendix_comparison}, we categorize state-of-the-art methods into VLMs-Based (subdivided into Supervised Fine-Tuning (SFT) and Reinforcement Fine-Tuning (RFT)) and RL-Based frameworks. Here, RL-Based specifically refers to non-VLM methods that utilize Reinforcement Learning for policy optimization.
As shown in~\cref{tab:appendix_comparison}, state-of-the-art methods are categorized into VLMs-Based and RL-Based frameworks, with VLMs-Based further divided into Supervised Fine-Tuning (SFT) and Reinforcement Fine-Tuning (RFT). RL-Based specifically refers to non-VLM approaches that leverage reinforcement learning for policy optimization.

%In the VLMs-Based (SFT) category, almost all models rely on billion-level parameters to encode linguistic and spatial priors. In contrast, UniTeD achieves competitive PDMS of 90.2 with only 6 decoder layers, demonstrating that our unified temporal diffusion paradigm effectively captures complex driving distributions without the prohibitive overhead of massive foundation models.

Within the VLMs-Based (SFT) category, most models depend on billion-level parameters to encode linguistic and spatial priors. In contrast, UniTeD achieves a competitive PDMS of 90.2 with only six decoder layers, demonstrating that our unified temporal diffusion paradigm effectively captures complex driving distributions without the prohibitive overhead of large foundation models.

%Within the VLMs-Based (RFT) and RL-Based categories, methods such as FLARE~\cite{zhang2025future} (PDMS 91.4) and DiffusionDriveV2~\cite{zou2025diffusiondrivev2} (PDMS 91.2) introduce reinforcement learning to obtain richer supervision and enhance closed-loop capabilities. Although UniTeD does not incorporate RL training, it still surpasses several RL-based models (e.g., AutoVLA~\cite{zhou2025autovla}, DriveDPO~\cite{shang2025drivedpo}, Trajhf~\cite{li2025finetuning}) and achieves highly competitive results against RFT methods.

Among VLMs-Based (RFT) and RL-Based methods, approaches such as FLARE~\cite{zhang2025future} (PDMS 91.4) and DiffusionDriveV2~\cite{zou2025diffusiondrivev2} (PDMS 91.2) incorporate reinforcement learning to obtain richer supervision and enhance closed-loop performance. Although UniTeD does not employ RL training, it still surpasses several RL-based models (e.g., AutoVLA~\cite{zhou2025autovla}, DriveDPO~\cite{shang2025drivedpo}, Trajhf~\cite{li2025learning}) and achieves results competitive with RFT methods.

In summary, UniTeD delivers strong performance with minimal model parameters and straightforward supervision. Our architecture is inherently scalable and can be readily extended to larger parameter sizes or integrated with reinforcement learning. We plan to explore these extensions in future work and anticipate that they will yield further performance gains.

% --- Table S2: 独立的浮动体 ---
\begin{table*}[!t]
\centering
\caption{Efficiency comparison on nuScenes val set. Latency of DriveTransformer is measured on H800 GPU, while others on RTX 4090 GPU.}
\label{tab:efficiency_full}
\resizebox{0.53\textwidth}{!}{
\begin{tabular}{@{}l|c|cc@{}}
    \toprule
    Method & Step & Latency (ms) $\downarrow$ & FPS $\uparrow$  \\
    \midrule
    UniAD~\cite{hu2023planning} & - & 555.6 & 1.8  \\
    VAD~\cite{jiang2023vad} & - & 222.2 & 4.5 \\
    DriveTransformer~\cite{jia2025drivetransformer} & - & 222.2 & 4.5  \\
    DiffusionDrive~\cite{liao2025diffusiondrive} & 2 & 122.0 & 8.2  \\
    SparseDrive~\cite{sun2025sparsedrive} & - & 111.1 & 9.0  \\
    HiP-AD~\cite{tang2025hip} & - & 109.9 & 9.1  \\
    \midrule
    \rowcolor{gray!15}
    \textbf{UniTeD} & 1 & 95.6 & 10.5 \\
    \rowcolor{gray!15}
    \textbf{UniTeD} & 2 & 155.3 & 6.4 \\
    \rowcolor{gray!15}
    \textbf{UniTeD} & 5 & 341.5 & 2.9 \\
    \bottomrule
\end{tabular}
}
\end{table*}

% \begin{table*}[!t]
% \centering
% % 增加 S3 和 S4 之间的物理间距，避免靠得太近

% \begin{floatrow}
%     % --- Table S3 ---
%     \ttabbox{%
%         \caption{Ablation of denoising steps on NAVSIM.}%
%         \label{tab:steps_abl}%
%     }{%
%         % 拉伸到 0.95\linewidth，使其占满半区（整体占满行宽）
%         \resizebox{0.6\linewidth}{!}{% 
%             \begin{tabular}{@{}c|c@{}}
%                 \toprule
%                 Steps & PDMS $\uparrow$ \\
%                 \midrule
%                 1 & 89.57 \\
%                 \rowcolor{gray!15} \textbf{2} & \textbf{90.23} \\
%                 5 & 89.95 \\
%                 % 【关键 Trick】：加一行隐形空行，强行对齐 S4 的高度
%                 \multicolumn{2}{c}{} \\ 
%                 \bottomrule
%             \end{tabular}%
%         }%
%     }
    
%     % --- Table S4 ---
%     \ttabbox{%
%         \caption{Sensitivity analysis of the $\tau$ (planning task) on NAVSIM.}%
%         \label{tab:threshold_abl}%
%     }{%
%         % 同样拉伸到 0.95\linewidth
%         \resizebox{0.8\linewidth}{!}{%
%             \begin{tabular}{@{}c|c@{}}
%                 \toprule
%                 $\tau$ (planning task) & PDMS $\uparrow$ \\
%                 \midrule
%                 0 & 88.23 \\
%                 0.2 & 89.72 \\
%                 \rowcolor{gray!15} \textbf{0.3} & \textbf{90.23} \\
%                 0.5 & 89.89 \\
%                 \bottomrule
%             \end{tabular}%
%         }%
%     }
% \end{floatrow}
% \end{table*}

\begin{table*}[!t]
\centering
\caption{Ablation of denoising steps on NAVSIM.}%
\label{tab:steps_abl}%
% 拉伸到 0.95\linewidth，使其占满半区（整体占满行宽）
%\resizebox{0.6\linewidth}{!}{% 
\begin{tabular}{@{}c|c@{}}
    \toprule
    Steps & PDMS $\uparrow$ \\
    \midrule
    1 & 89.57 \\
    \rowcolor{gray!15} \textbf{2} & \textbf{90.23} \\
    5 & 89.95 \\
    % 【关键 Trick】：加一行隐形空行，强行对齐 S4 的高度
    \multicolumn{2}{c}{} \\ 
    \bottomrule
\end{tabular}%
%}%
\end{table*}

\begin{table*}[!t]
\centering
\caption{Sensitivity analysis of the $\tau$ (planning task) on NAVSIM.}%
\label{tab:threshold_abl}%
%
        % 同样拉伸到 0.95\linewidth
%        \resizebox{0.8\linewidth}{!}{%
\begin{tabular}{@{}c|c@{}}
    \toprule
    $\tau$ (planning task) & PDMS $\uparrow$ \\
    \midrule
    0 & 88.23 \\
    0.2 & 89.72 \\
    \rowcolor{gray!15} \textbf{0.3} & \textbf{90.23} \\
    0.5 & 89.89 \\
    \bottomrule
\end{tabular}%
\end{table*}

\subsection{Inference Efficiency Analysis}
\label{app:efficiency_analysis}
Following standard evaluation protocols, we comprehensively assess the inference efficiency of our method on the nuScenes validation set, as detailed in ~\cref{tab:efficiency_full}. While our default configuration utilizes 2 denoising steps to achieve optimal performance, UniTeD is highly flexible and can operate with just 1 denoising step. In this 1-step setting, UniTeD achieves an impressive inference speed of 10.5 FPS (95.6 ms latency) on a single RTX 4090 GPU. This demonstrates a substantial speed advantage over existing state-of-the-art methods, including UniAD~\cite{hu2023planning} (1.8 FPS) and VAD~\cite{jiang2023vad} (4.5 FPS), and even outperforms highly optimized sparse-query baselines such as SparseDrive~\cite{sun2025sparsedrive} (9.0 FPS).

Furthermore, we investigate the impact of denoising steps on the planning performance using the NAVSIM benchmark, as shown in ~\cref{tab:steps_abl}. Although the default 2-step inference yields the optimal PDMS of 90.23, reducing the steps to 1 maintains highly competitive accuracy at 89.57 PDMS. These results confirm that UniTeD provides a highly practical and deployable solution, effectively balancing state-of-the-art planning accuracy with real-time inference efficiency.

\subsection{Details of Anchor Refresh Strategy (ARS)}
\label{app:ars_details}
In our diffusion process, the confidence scores for the ARS are directly derived from the classification branch, serving as a quantitative measure of the model's certainty for each anchor. During the iterative denoising steps, predictions yielding a classification confidence above a predefined threshold $\tau$ are retained for subsequent refinement. Conversely, predictions falling below $\tau$ are discarded and revert to their original anchor priors.

To evaluate the impact of this threshold, we conduct a sensitivity analysis on the planning task, as shown in ~\cref{tab:threshold_abl}. Setting $\tau=0$ effectively disables the ARS mechanism, allowing all predictions, including low-confidence noisy ones, to propagate to the next step. As $\tau$ increases to 0.3, the planning performance improves significantly. This indicates that filtering out uncertain predictions successfully aligns the iterative inputs more closely with the training data distribution, establishing $\tau=0.3$ as the optimal empirical setting. However, applying an excessively high threshold (\textit{e.g.}, $\tau=0.5$) leads to the over-pruning of high-quality refined predictions. This forces the model to revert too frequently to the raw anchor priors, resulting in a slight degradation in overall accuracy.

\section{Vistualization}
\label{sec:vis}
\subsection{Visualizations on NAVSIM}
%We provide extensive visualizations on NAVSIM~\cite{2024NAVSIM} to compare UniTeD with the unified non-diffusion baseline. In standard scenarios, both UniTeD and the baseline achieve comparable performance. To clearly highlight the advantages of our approach, we focus on relatively complex scenarios in this section. 

%Visual annotations: Yellow circles highlight visual cues in the input. Red circles indicate baseline errors. Green circles show correct predictions from UniTeD and GT.

We provide extensive qualitative comparisons between UniTeD and a unified non-diffusion baseline on the NAVSIM benchmark~\cite{2024NAVSIM}. While both methods perform comparably in standard scenarios, we focus here on more complex situations to better highlight the advantages of our approach. In the presented visualizations, yellow circles highlight key visual cues in the input, red circles indicate errors made by the baseline, and green circles denote correct predictions from UniTeD and the ground truth (GT).

\subsubsection{Robustness via Generative Uncertainty Modeling}

\begin{figure}[!t]
    \centering
    \begin{subfigure}[b]{\linewidth}
        \includegraphics[width=\linewidth]{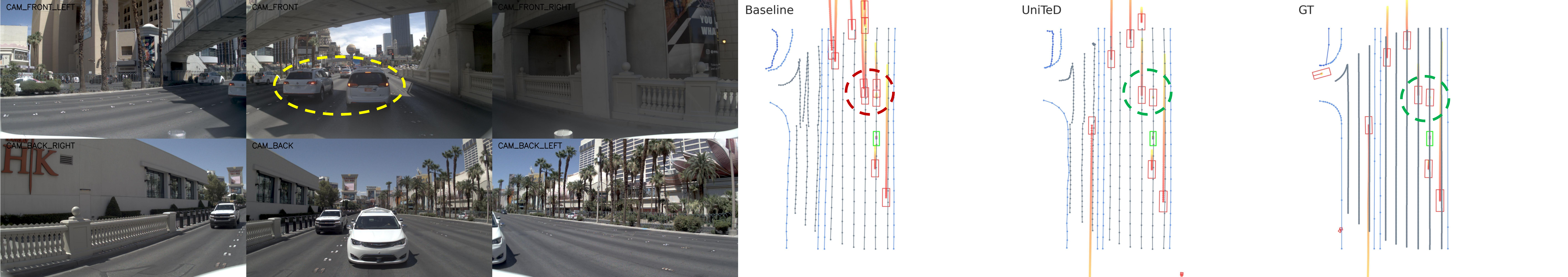}
        \subcaption{Shadows}
        \label{fig:sfig1_a}
    \end{subfigure}
    
    \begin{subfigure}[b]{\linewidth}
        \includegraphics[width=\linewidth]{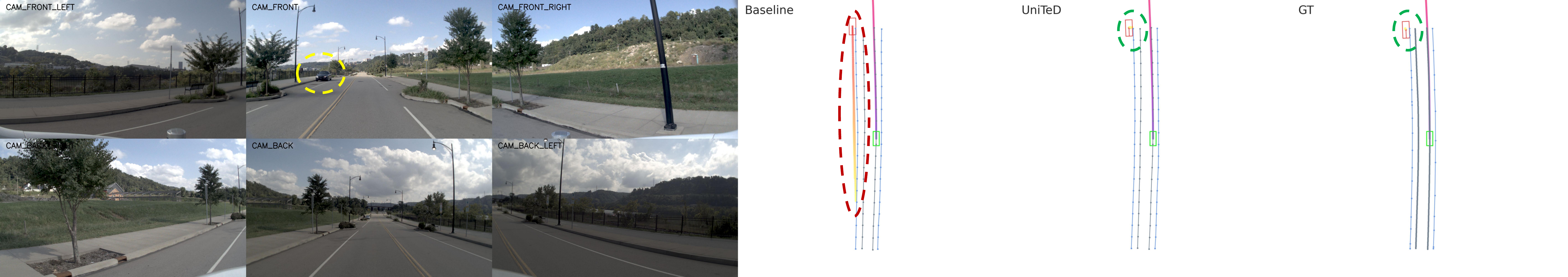}
        \subcaption{Parked agent}
        \label{fig:sfig1_b}
    \end{subfigure}
        
    \begin{subfigure}[b]{\linewidth}
        \includegraphics[width=\linewidth]{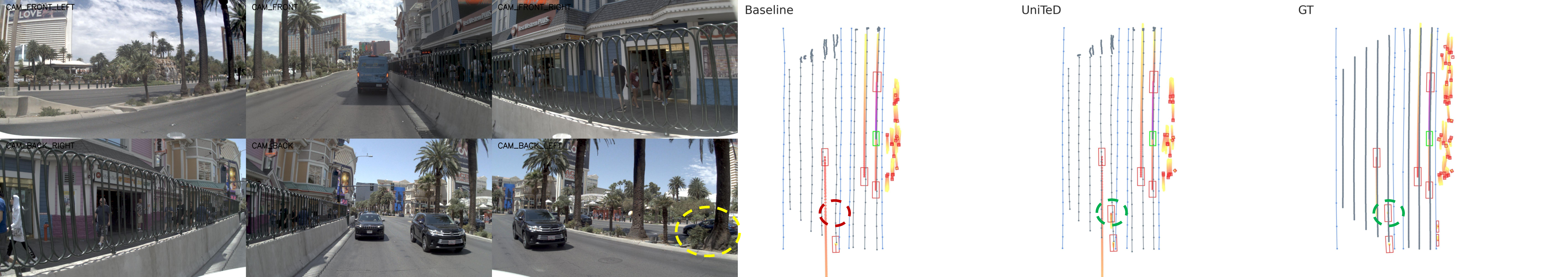}
        \subcaption{Occluded agent}
        \label{fig:sfig1_c}
    \end{subfigure}
        
    \begin{subfigure}[b]{\linewidth}
        \includegraphics[width=\linewidth]{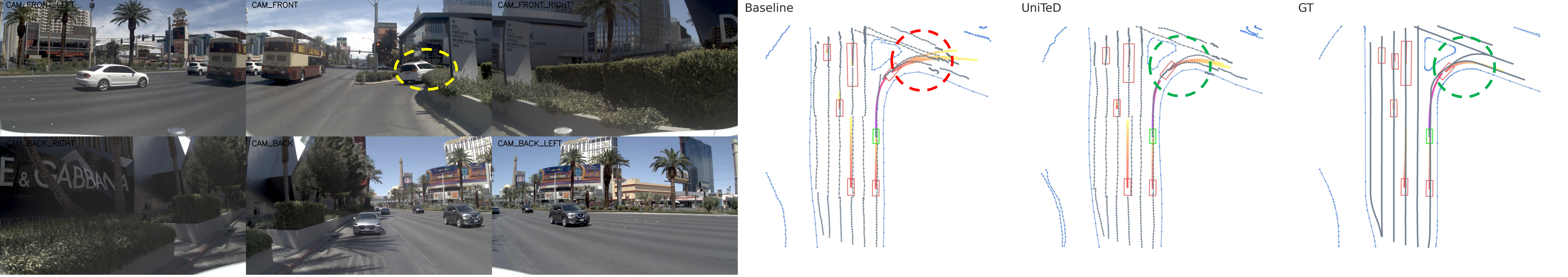}
        \subcaption{Occluded turn}
        \label{fig:sfig1_d}
    \end{subfigure}
    
    \caption{Qualitative Comparison on NAVSIM: Robustness via Generative Uncertainty Modeling.} 
    \label{fig:s_figure1}
\end{figure}

The generative paradigm offers inherent advantages in handling input noise and ambiguous scenes. Non-diffusion methods rely on deterministic regression, which generates a fixed deterministic output and struggles when visual cues are corrupted or insufficient. In contrast, diffusion models naturally capture multi-modal distributions through iterative denoising, enabling robust predictions by reasoning over multiple plausible hypotheses.
~\cref{fig:s_figure1} illustrates this capability across four representative scenarios.
%Fig.~\cref{fig:s_figure1} illustrates four representative scenarios.

\noindent\textbf{(a) Shadows.} 
As shown in~\cref{fig:sfig1_a}, under heavy shadows, the baseline produces false positives due to sensitivity to visual noise. 
In contrast, UniTeD demonstrates inherent robustness to environmental perturbations, maintaining accurate perception even under challenging lighting conditions.
%UniTeD is inherently robust to environmental noise, maintaining accurate perception under adverse lighting conditions.

\noindent\textbf{(b) Parked agent.} 
As shown in~\cref{fig:sfig1_b}, the baseline erroneously predicts motion for a stationary agent. This occurs because deterministic regression tends to output the statistical average of the training data (i.e., most agents are moving) rather than reasoning about the specific scene context. By evaluating multiple plausible hypotheses, UniTeD correctly identifies the static status, selecting the prediction most consistent with the observed context.

%As shown in~\cref{fig:sfig1_b}, the baseline predicts false motion for stationary agents, as deterministic regression tends to output the statistical average of the training data (i.e., most agents are moving) rather than reasoning about the specific scene context. UniTeD correctly identifies the static status by evaluating multiple plausible hypotheses and selecting the one most consistent with the observed context.

\noindent\textbf{(c) Occluded agent.} 
As shown in~\cref{fig:sfig1_c}, when agents are partially occluded, the baseline misses detections due to incomplete visual features. 
UniTeD, however, robustly recovers the occluded object by reasoning over its potential states, leveraging the generative model's capacity to infer under uncertainty.
%UniTeD is inherently robust to noisy conditions, effectively recovering occluded objects by reasoning over multiple plausible states.

\noindent\textbf{(d) Occluded turn.} 
As shown in~\cref{fig:sfig1_d}, under partial observability at an intersection, the baseline predicts an incorrect topology and generates motion that contradicts the map structure. This failure stems from its inability to capture multi-modal possibilities. By sampling from the joint distribution of motion and topology, UniTeD generates a reasonable map and ensures the predicted trajectory remains self-consistent with it.

%As shown in~\cref{fig:sfig1_d}, under partial observability at turns, the baseline predicts incorrect topology and simultaneously produces erroneous motion that contradicts the map structure, as deterministic regression cannot capture multi-modal possibilities. UniTeD generates reasonable topology and ensures the motion remains self-consistent with the generated map structure through joint distribution sampling.

These results demonstrate that UniTeD leverages generative uncertainty modeling to achieve robust result under noise, occlusion, and ambiguity, which addresses key limitations of deterministic regression approaches.

\subsubsection{Mutual Refinement and Structural Consistency via Joint Distribution Modeling}
\begin{figure}[!t]
    \centering
    \begin{subfigure}[b]{\linewidth}
        \includegraphics[width=\linewidth]{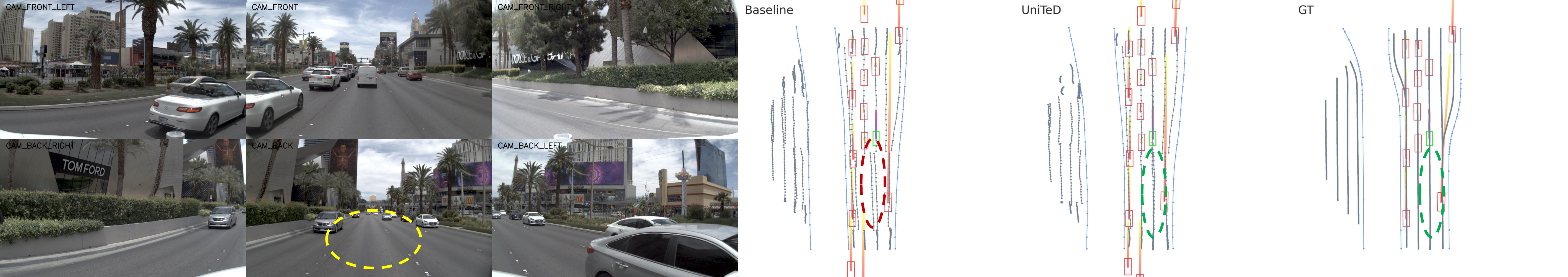}
        \subcaption{Ambiguous lanes}
        \label{fig:sfig2_a}
    \end{subfigure}
        
    \begin{subfigure}[b]{\linewidth}
        \includegraphics[width=\linewidth]{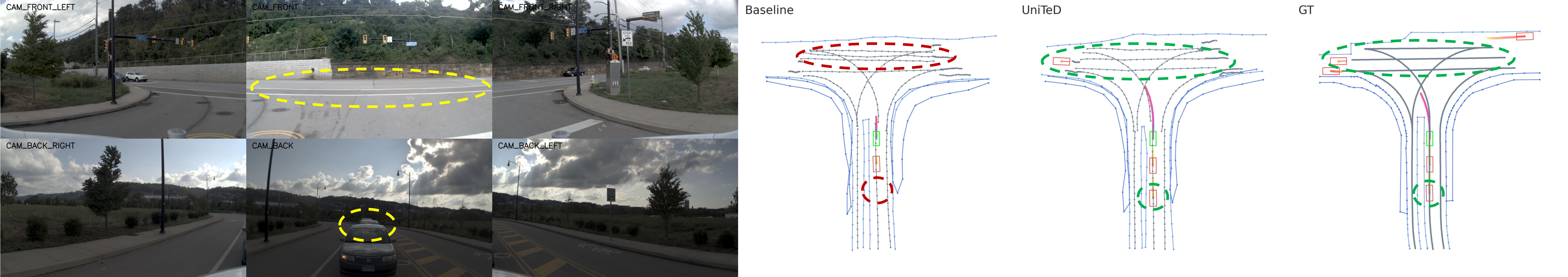}
        \subcaption{Ambiguous T-junction}
        \label{fig:sfig2_b}
    \end{subfigure}
        
    \begin{subfigure}[b]{\linewidth}
        \includegraphics[width=\linewidth]{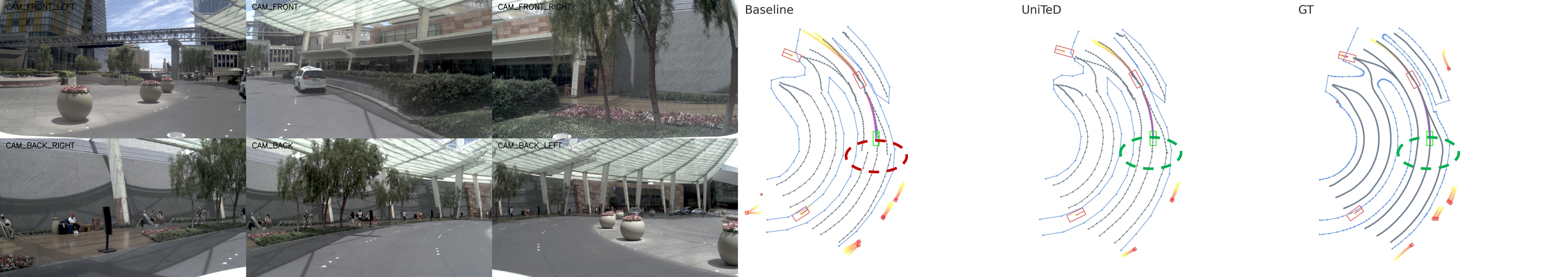}
        \subcaption{Sharp curve with merge}
        \label{fig:sfig2_c}
    \end{subfigure}
        
    \begin{subfigure}[b]{\linewidth}
        \includegraphics[width=\linewidth]{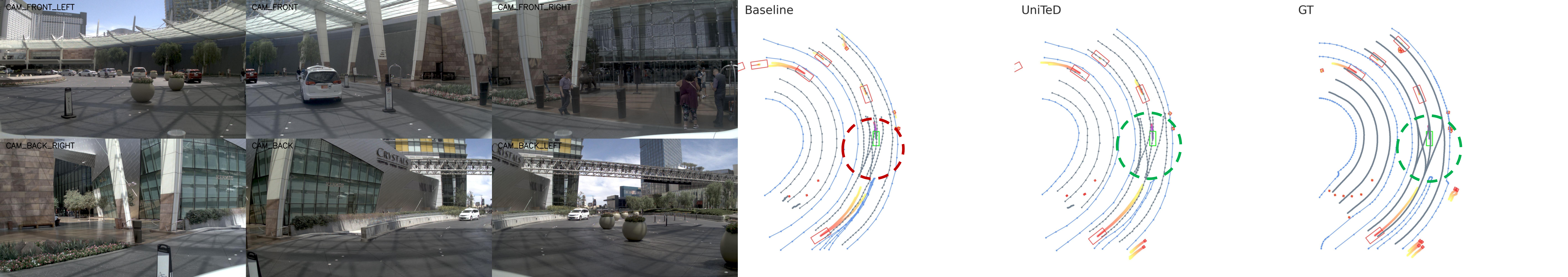}
        \subcaption{Sharp curve with split}
        \label{fig:sfig2_d}
    \end{subfigure}
    
    \caption{Qualitative Comparison on NAVSIM: Mutual Refinement and Structural Consistency via Joint Distribution Modeling.} 
    \label{fig:s_figure2}
\end{figure}

Beyond uncertainty modeling, sampling from a multi-task joint distribution enables mutual refinement between different tasks, ensuring structural consistency across outputs. Unlike the decoupled optimization of baseline, UniTeD inherently correlates map, agent, and planning within a unified generative process.~\cref{fig:s_figure2} illustrates this capability across four representative scenarios.

%illustrates this capability across four representative scenarios.

\noindent\textbf{(a) Ambiguous lanes.} 
As shown in~\cref{fig:sfig2_a}, when lane markings are faint or ambiguous, the baseline produces erroneous centerline predictions due to reliance on local visual cues. In contrast, UniTeD leverages the motion patterns of the ego vehicle and surrounding agents to infer the correct lane structure, demonstrating cross-task refinement through joint distribution modeling.

\noindent\textbf{(b) Ambiguous T-junction.} 
As shown in~\cref{fig:sfig2_b}, at an ambiguous T-junction, the baseline produces incorrect map topology and misses several agents. This perception failure causes the ego vehicle to misinterpret the scene structure, resulting in overly conservative planning with significantly reduced speed. By jointly sampling map, agent, and planning queries, UniTeD enables mutual refinement: agent motion helps infer correct topology, accurate topology constrains plausible agent trajectories, and coherent planning aligns with both. This holistic generation yields consistent map-agent-planning predictions. Although a small distant agent is missed, the overall scene understanding remains accurate and does not affect the ego vehicle's final decision. Notably, the occluded agent that the baseline fails to detect is accurately identified by UniTeD.

%As shown in~\cref{fig:sfig2_b}, at ambiguous T-junctions, the baseline produces incorrect map topology and misses agents, and the perception failure causes the ego vehicle to become confused about the scene structure, resulting in overly conservative planning with significantly reduced speed. UniTeD jointly samples map, agent, and planning queries, enabling mutual refinement: agent motion helps infer correct topology, accurate topology constrains plausible agent trajectories, and coherent planning aligns with both, resulting in consistent map-agent-planning predictions. Although a small distant agent is missed, the overall scene understanding remains accurate and does not affect the ego vehicle's final decision. By the way, the occluded agent that the baseline fails to detect is accurately identified by UniTeD. 

\noindent\textbf{(c) Sharp curve with merge.} 
As shown in~\cref{fig:sfig2_c}, at a merge point on a sharp curve, the baseline exhibits centerline discontinuities due to independent regression of each map element. UniTeD jointly samples all tasks through the generative process, producing smooth and continuous centerlines with accurate topology across the merge region.

\noindent\textbf{(d) Sharp curve with split.} 
As shown in~\cref{fig:sfig2_d}, in a complex split scenario, the baseline predicts broken or misaligned centerlines due to decoupled optimization, and the associated agent motion also deviate from the correct path. 
Through joint sampling, UniTeD ensures centerlines are correctly connected, geometrically smooth, and topologically precise, with agent motions consistently aligned to the accurate topology as part of a holistic scene generation.
%UniTeD jointly samples all tasks through the generative process, ensuring centerlines are correctly connected, geometrically smooth, and topologically precise, with agent motions consistently aligned to the accurate topology through holistic scene generation.

These results demonstrate that UniTeD leverages joint distribution modeling to enable mutual refinement among map, agent, and planning while ensuring structural consistency, effectively addressing key limitations of decoupled regression approaches.

%======================================
\subsection{Visualizations on Bench2Drive}
\begin{figure}[!t]
    \centering
    \includegraphics[width=1\linewidth]{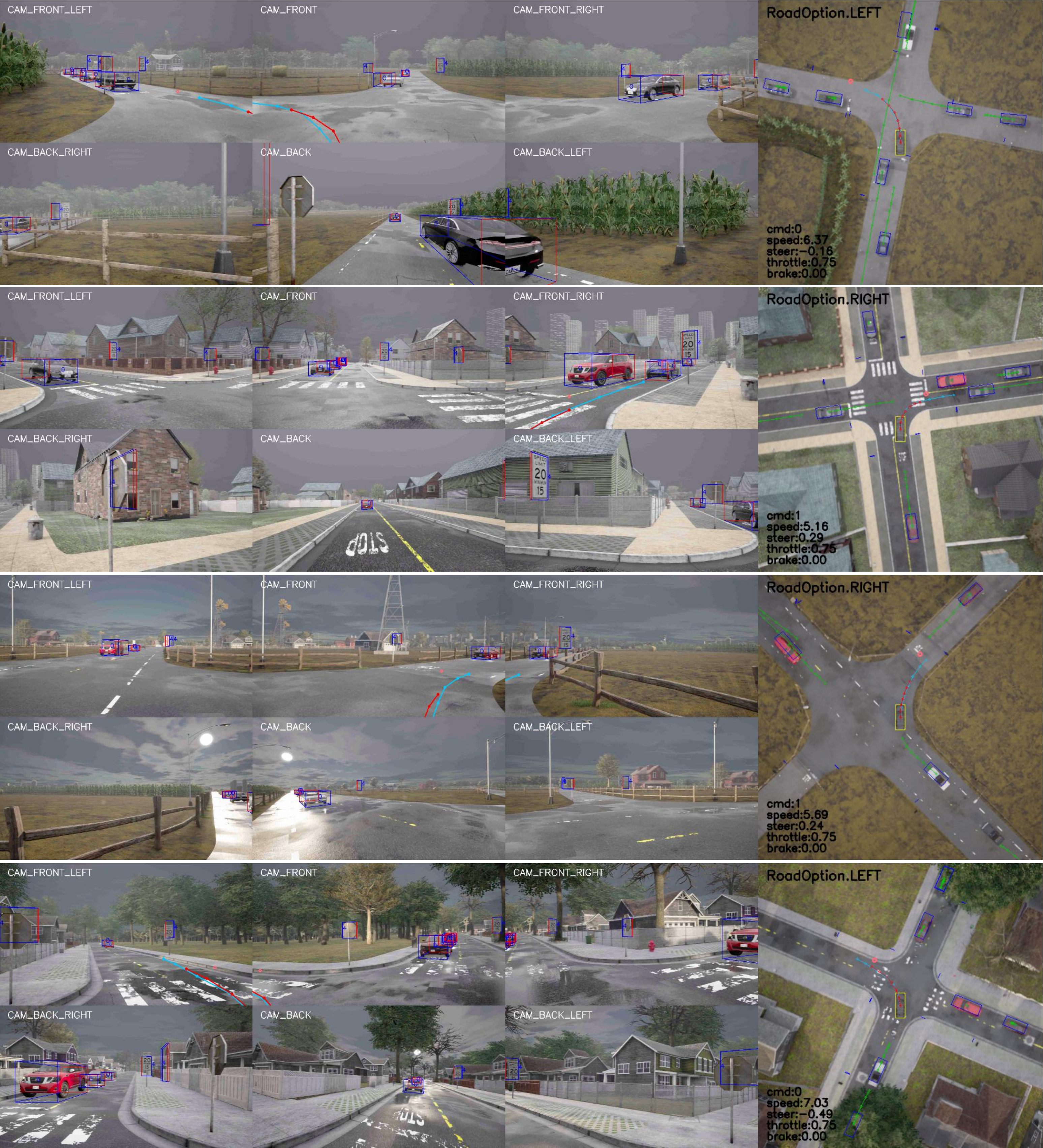}
    \caption{UniTeD's Closed-Loop Planning on Bench2Drive: Cross.}
    \label{fig:cross}
\end{figure}
\begin{figure}[!t]
    \centering
    \includegraphics[width=1\linewidth]{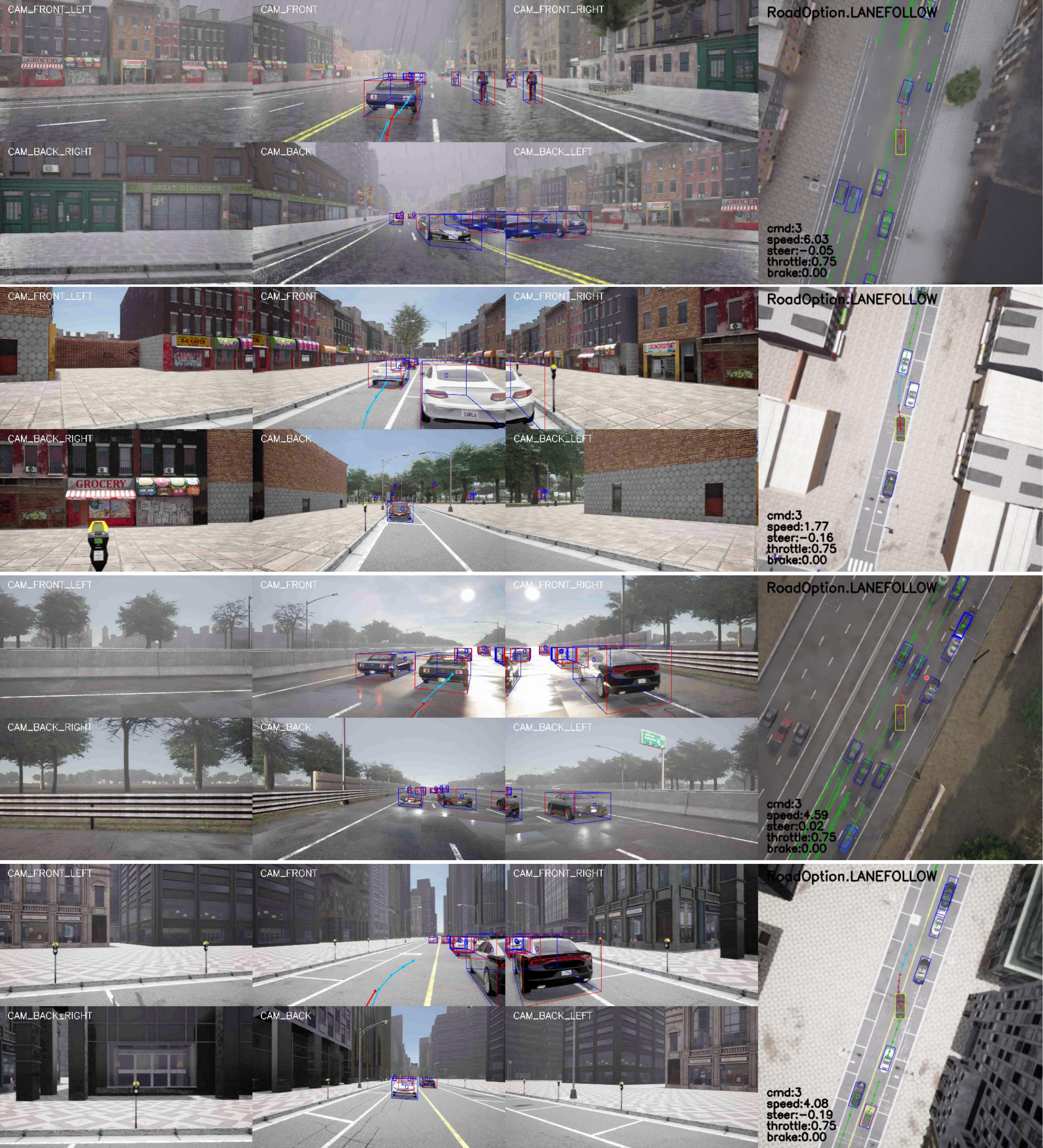}
    \caption{UniTeD's Closed-Loop Planning on Bench2Drive: Overtaking.}
    \label{fig:overtaking}
\end{figure}
\begin{figure}[!t]
    \centering
    \includegraphics[width=1\linewidth]{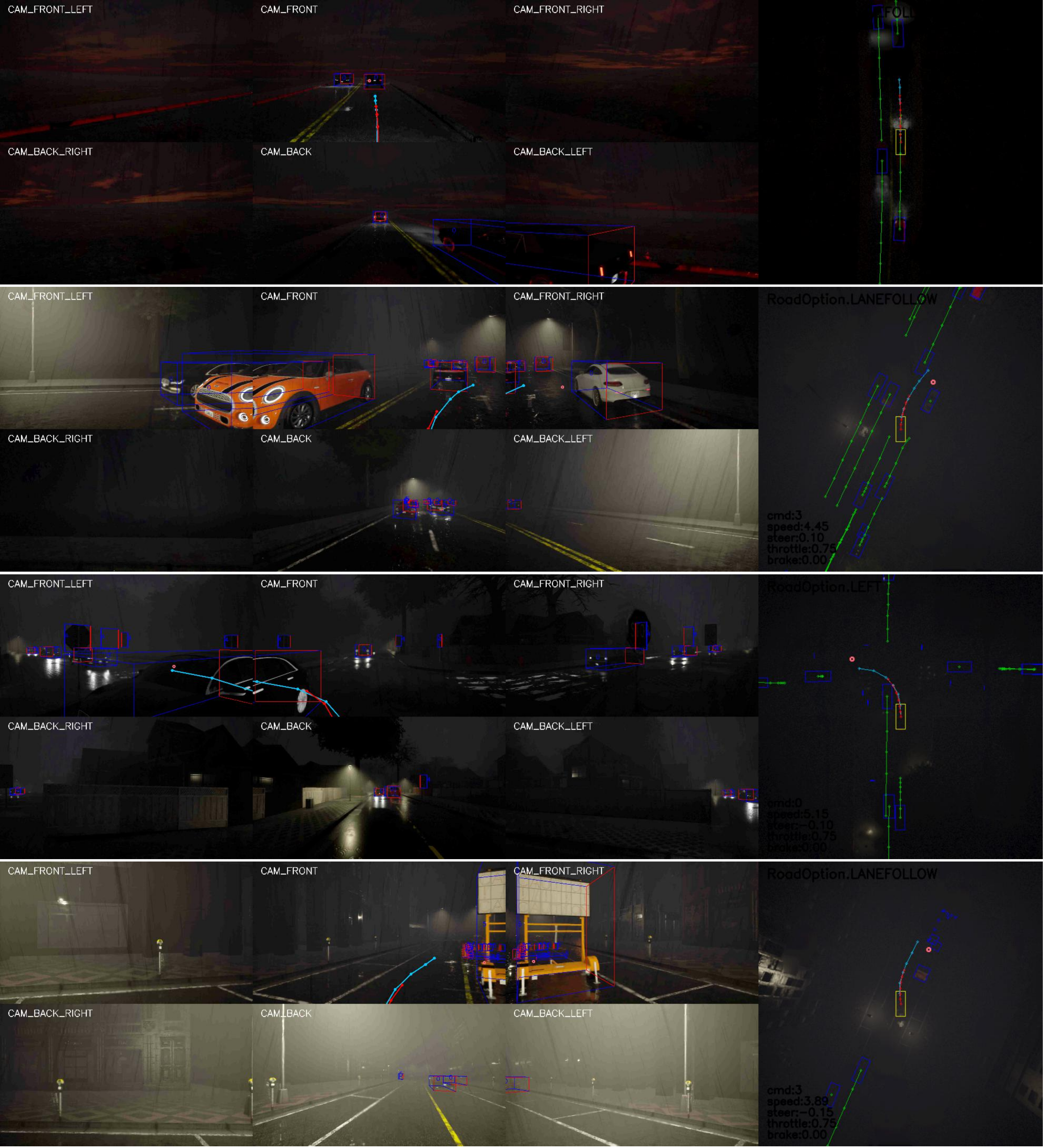}
    \caption{UniTeD's Closed-Loop Planning on Bench2Drive: Night.}
    \label{fig:night}
\end{figure}

We provide qualitative results of UniTeD on the Bench2Drive closed-loop benchmark to demonstrate its driving capability in dynamic environments. Bench2Drive evaluates E2E driving performance in interactive scenarios, not only perception errors can propagate to planning failures, but historical planning errors are also infinitely accumulated and amplified in closed-loop systems over time. This error accumulation is a fundamental challenge for most E2E driving models. UniTeD's generative paradigm is inherently robust to noise through multi-modal distribution modeling. Crucially, at each control cycle, UniTeD regenerates predictions from the learned joint distribution conditioned on the current observation. This noise-conditioned generation allows UniTeD to correct historical errors at every iteration, effectively resetting accumulated noise and preventing error propagation throughout the closed-loop execution.

Unlike the NAVSIM experiments where UniTeD outputs a unified trajectory, on Bench2Drive we adopt a decoupled lateral-longitudinal planning formulation for more precise control: the model predicts a spatial path (lateral) and a temporal trajectory (longitudinal) separately. In the following figures, blue lines represent the predicted path for the next 20 meters (lateral planning), while red lines represent the predicted trajectory for the next 3 seconds (longitudinal planning).

As shown in~\cref{fig:cross}, UniTeD demonstrates superior performance in complex intersection scenarios. The ego vehicle approaches a busy intersection with crossing traffic and ambiguous right-of-way. UniTeD generates smooth, socially compliant trajectories by jointly reasoning about map topology, agent intentions, and ego planning in a unified generative process. The model correctly infers the drivable path through the intersection, maintains safe distances with crossing agents, and produces kinematically feasible acceleration profiles without oscillations.

As shown in~\cref{fig:overtaking}, UniTeD exhibits reasonable and smooth overtaking behaviors in dynamic traffic. The ego vehicle needs to overtake a slow-moving or parked vehicle while considering oncoming traffic and lane constraints. UniTeD generates smooth, kinematically feasible overtaking trajectories by jointly sampling map, agent, and planning queries. The model accurately assesses the gap in oncoming traffic, initiates lane change at the appropriate timing, and completes the maneuver with continuous curvature and comfortable acceleration, ensuring both safety and traffic efficiency.

As shown in~\cref{fig:night}, UniTeD maintains reliable planning performance under challenging low-light conditions. In night driving scenarios with reduced visibility, lane markings and agent appearances are less distinct. UniTeD leverages generative uncertainty modeling to maintain robust prediction under low-light conditions. The model effectively recovers lane structures from faint markings, infers agent states despite reduced visual clarity, and enables stable, safe planning decisions through joint distribution sampling. The resulting planning remains centered in the lane with appropriate speed adaptation to visibility conditions.

The visualizations reveal that UniTeD maintains safe and rational driving behavior even in interactive closed-loop settings. By jointly sampling map, agent, and planning queries from the learned distribution, the model accurately perceives map structures and surrounding agents, and generates collision-free, kinematically feasible planning outputs. Crucially, the noise-conditioned generation mechanism corrects historical errors at each control cycle, preventing error accumulation. This demonstrates that our unified generative approach achieves robust scene understanding and human-like driving performance in E2E autonomous navigation.

\end{document}